\pdfoutput=1
\documentclass[10pt]{article}

\usepackage[margin=1in]{geometry}
\usepackage[T1]{fontenc}
\usepackage[utf8]{inputenc}
\usepackage{lmodern}
\usepackage{microtype}
\usepackage{amsmath,amssymb}

\usepackage{booktabs}
\usepackage{array}
\usepackage{multirow}
\usepackage[table]{xcolor}
\usepackage{graphicx}
\usepackage{subcaption}
\usepackage{tikz}
\usetikzlibrary{positioning,arrows.meta,shapes.geometric}
\usepackage{pgfplots}
\pgfplotsset{compat=1.18}
\usepackage[
  backend=biber,
  style=apa,
  natbib=true,
  giveninits=true,
  uniquename=false
]{biblatex}
\usepackage[hidelinks]{hyperref}
\usepackage{enumitem}
\usepackage[ruled,vlined,linesnumbered]{algorithm2e}
\SetKwInput{KwIn}{Input}
\SetKwInput{KwOut}{Output}
\usepackage{authblk}
\usepackage{verbatim}
\usepackage{fvextra}
\usepackage{dirtytalk}
\usepackage{placeins}
\usepackage{float}

\DefineVerbatimEnvironment{SystemPrompt}{Verbatim}{
  fontsize=\footnotesize,
  breaklines=true,
  breakanywhere=true,
  breaksymbolleft={},
  breaksymbolright={},
  frame=single,
  framerule=0.4pt,
  framesep=2mm,
  rulecolor=\color{black!55},
  label={\sffamily\bfseries System message},
  labelposition=topline
}
\DefineVerbatimEnvironment{HumanPrompt}{Verbatim}{
  fontsize=\footnotesize,
  breaklines=true,
  breakanywhere=true,
  breaksymbolleft={},
  breaksymbolright={},
  frame=single,
  framerule=0.4pt,
  framesep=2mm,
  rulecolor=\color{black!55},
  label={\strut\textsf{\textbf{Human message}}},
  labelposition=topline
}
\DefineVerbatimEnvironment{AIPrompt}{Verbatim}{
  fontsize=\footnotesize,
  breaklines=true,
  breakanywhere=true,
  breaksymbolleft={},
  breaksymbolright={},
  frame=single,
  framerule=0.4pt,
  framesep=2mm,
  rulecolor=\color{black!55},
  label={\strut\textsf{\textbf{AI message}}},
  labelposition=topline
}
\DefineVerbatimEnvironment{UserGuide}{Verbatim}{
  fontsize=\footnotesize,
  breaklines=true,
  breakanywhere=true,
  breaksymbolleft={},
  breaksymbolright={},
  frame=single,
  framerule=0.4pt,
  framesep=2mm,
  rulecolor=\color{black!55}
}

\title{ParetoChat: An LLM-Based Interface for Interactive Multi-Objective Optimization}
\author{G\"{o}khan Ceyhan}
\affil{Just Eat Takeaway.com}

\begin{document}
\maketitle

\begin{abstract}
Multi-objective optimization models are often deployed in settings where decision makers (DMs) must balance some conflicting criteria. In practice, however, stakeholders must translate their qualitative preferences between different criteria into numerical weights, priorities, or reference points required by optimization methods, which can be difficult to specify and interpret. This paper proposes \emph{ParetoChat}, an LLM (large language model)-based expert-agent interface for interactive multi-objective optimization. The agent translates natural-language user input into structured scalarization parameters, invokes an optimization system, presents candidate solutions, and iteratively elicits feedback. We describe a multi-agent architecture that separates input processing, multi-criteria decision-making (MCDM) expertise, solution presentation, and feedback processing. We then present a simulation study using an LLM-based DM agent and benchmark ParetoChat against interactive methods implemented in DESDEO, including reference-point and NIMBUS methods. The computational experiments on multi-objective binary knapsack and set-partitioning instances show that ParetoChat can guide the search efficiently toward solutions that match the DM's preferences. An application to historical courier-order assignment instances illustrates how users can explore objective extremes and intermediate trade-offs through natural-language requests. The results indicate that LLM agents may provide a practical interaction layer between business stakeholders and optimization tools.
\end{abstract}

\section{Introduction}
Many operational decision problems are not naturally single-objective. Delivery platforms, logistics systems, workforce planning tools, and pricing engines must typically trade off multiple objectives such as cost, service quality, and reliability. At Just Eat Takeaway (JET), examples include real-time capacity management, courier-order assignment, and dynamic courier-offer pricing. In these applications, stakeholders rarely care only about optimizing a single objective. Instead, they want to understand what trade-offs are possible, identify practically acceptable compromise solutions, and steer the system as business priorities change.

A common practical approach is to convert the multi-objective problem into a single optimization problem before optimization begins. For instance, objectives may be combined using a weighted-sum. This \emph{a priori} scalarization approach is convenient for solvers, but it creates a usability gap. DMs often express preferences in qualitative and contextual language, while optimization systems expect precise numerical parameters. Even when the DM is willing to provide numerical weights, those weights can be difficult to interpret because objective scales, units, and nonlinear trade-offs affect the realized decision. As a result, tuning scalarization parameters may require many expensive iterations, yet the resulting solution may still fail to reflect the DM's actual preferences.

Interactive multi-objective optimization addresses this gap by involving the DM during the search process. Rather than fixing all preferences at the beginning, the system presents candidate solutions, requests feedback, and refines the search accordingly. A central challenge remains: classical interactive methods still require the DM to provide structured preference information, such as aspiration levels, classification of objectives, or pairwise trade-off information. These inputs can be unintuitive for non-expert stakeholders, especially when they are using the method through an operational decision-support tool rather than as optimization specialists. Some methods also impose assumptions on the form of the DM's value function, which may limit their ability to represent the DM's actual preferences. This paper explores the use of LLM agents as an interaction layer for multi-objective optimization. The proposed system, called \emph{ParetoChat}, acts as an expert interface that translates natural-language preferences into solver-compatible parameters, calls existing optimization tools, explains solutions and trade-offs, and maintains conversational context. The user remains in control in a co-pilot interaction mode: they can ask for new solutions, save or reject candidates, summarize the session, or refine their preferences in natural language.

This paper makes three main contributions. First, we propose a conversational approach to interactive multi-objective optimization in which an LLM-based multi-agent system translates qualitative preferences and feedback into successive scalarized optimization problems. This approach allows the DM to steer the search through natural language without directly specifying numerical scalarization parameters, while keeping preference interpretation separate from mathematical optimization. Second, we introduce an evaluation framework that combines an LLM-based artificial DM with a hidden value function: the agent interacts with each method according to its supplied user guide and generates adaptive requests, while the value function provides a consistent basis for measuring solution quality. This framework enables comparisons across different interactive multi-objective optimization methods using different interaction formats with DMs. Third, we provide empirical evidence of the capabilities and limitations of conversational preference elicitation. The simulations with ParetoChat reveal a trade-off between fewer solution evaluations and less consistent solution quality, with additional iterations improving results in the tested preference settings. The courier-order assignment application complements these findings by demonstrating how conversational references to earlier solutions support trade-off exploration, while exposing preference-translation errors that motivate transparent scalarization parameters and opportunities for user correction.

The remainder of the paper is organized as follows. Section~\ref{sec:background} reviews multi-objective optimization, interactive methods, and related applications of LLMs in optimization. Section~\ref{sec:paretochat} presents the ParetoChat architecture and an illustrative interaction. Section~\ref{sec:simulation} describes the simulation framework, benchmark problems, evaluation metrics, and comparative results. Section~\ref{sec:case-study} presents a real-life case study and applies ParetoChat to some historical courier-order assignment instances from JET's dispatch system. Section~\ref{sec:discussion} discusses the findings, behavioral considerations, and limitations, and Section~\ref{sec:conclusions} concludes the paper and outlines directions for future work.

\section{Background and Related Work}
\label{sec:background}
\subsection{Multi-objective optimization}
A multi-objective optimization problem (MOOP) can be written as
\begin{equation}
\label{eq:moop}
  \max_{x \in X} \; z(x) = \bigl(z_1(x), z_2(x), \ldots, z_m(x)\bigr),
\end{equation}
where $X$ denotes the feasible set and $z_i(x)$ denotes the value of objective $i$. Because objectives often conflict, there is generally no single solution that optimizes all objectives simultaneously. The relevant output is therefore a set of nondominated points or Pareto-efficient solutions, each representing a different trade-off.

Formally, for a maximization problem, a feasible solution $x^* \in X$ is called an \emph{efficient solution} (or Pareto-efficient solution) if there is no other feasible solution $x \in X$ such that
\begin{equation}
  z_i(x) \ge z_i(x^*) \quad \text{for all } i=1,\ldots,m,
  \quad \text{and} \quad
  z_j(x) > z_j(x^*) \text{ for at least one } j.
\end{equation}
Let $Z$ represent the feasible objective space, $Z=\left\{z(x): x \in X \right\}$. Assume $Z$ is compact. The objective vector $z(x^*) \in Z$ corresponding to an efficient solution is called a \emph{nondominated point} (A solution $x^*$ is called weakly efficient and $z(x^*)$ is called weakly nondominated if $\nexists x \in X: z(x_i) > z(x^*_i), \forall i=1,\ldots,m$). Equivalently, a nondominated point is an attainable objective vector for which no other attainable objective vector improves at least one objective without worsening any other objective. Note that there might exist many such solutions or points for a given MOOP. We denote the set of efficient solutions as $X^E \subseteq X$ and the set of nondominated points as $Z^{ND} \subseteq Z$.

There are also two special points for MOOPs, the ideal and the nadir points. Let $z^{I}$ denote the ideal point, where $z^{I}_i=\max_{x\in X} z_i(x)$, and $z^{N}$ denote the nadir point, where $z^{N}_i=\min_{x\in X^E} z_i(x)$. Note that any nondominated point lies in the hyperrectangle defined by these two points, that is, $z^{N}_i \leq z_i(x) \leq z^{I}_i, \forall i=1,\ldots,m, \forall x \in X^E$.

In many business settings, however, DMs are looking for a single solution. A common strategy is to transform the vector optimization problem into a scalar optimization problem via scalarization techniques. Some well-known scalarization approaches include weighted-sum, lexicographic optimization, achievement scalarizing functions, and the $\epsilon$-constraint method (see Appendix~\ref{app:scalarization-methods}). 



\subsection{Interactive multi-objective optimization}
Interactive methods involve the DM in the optimization process. Instead of generating an entire Pareto front upfront, these methods iteratively propose solutions and obtain preference information from the DM to focus the search on regions of interest. This is especially useful when preferences are not fully known before optimization, when the nondominated set is too large to inspect, or when the DM learns about feasible trade-offs during the solution process \citep{miettinen1999nonlinear,miettinen2008introduction,alves2007review}.

Early interactive multi-objective optimization methods were developed for mathematical programming settings and differ mainly in the type of preference information requested from the DM. The Geoffrion--Dyer--Feinberg method uses local trade-off information to estimate a value function and move toward more preferred solutions \citep{geoffrion1972interactive}. STEM asks the DM to relax one or more objectives after seeing a compromise solution, thereby progressively reducing the feasible region \citep{benayoun1971linear}. Zionts--Wallenius uses pairwise preference judgments to infer compatible weights for a linear additive value function and generate improved alternatives \citep{zionts1976interactive}. The interactive weighted Tchebycheff procedure (IWTP) of \citet{steuer1983interactive} presents the DM with a sample of nondominated points generated using weighted Tchebycheff scalarizations, then progressively narrows the sampled weight space according to the point selected as most preferred.

Reference-point and achievement-scalarizing-function methods form another central family. In these methods, the DM specifies aspiration or reservation levels, and the method solves scalarized subproblems to find nondominated points close to the indicated direction or target \citep{wierzbicki1982mathematical,miettinen2008introduction}. These methods are attractive in decision-support contexts because aspiration levels are often easier to interpret than abstract weights. They have also been adapted to multi-objective integer and mixed-integer programming, where exact generation of the whole nondominated set can be computationally demanding \citep{alves2000interactive,alves2007review}.

Classification-based methods ask the DM to classify objectives according to desired changes. The NIMBUS method is a well-known representative: after inspecting a current solution, the DM classifies each objective as one to improve, improve until an aspiration level, keep unchanged, allow to deteriorate, or change freely; these classifications are then translated into scalarized subproblems that generate new candidate solutions \citep{miettinen2006synchronous}. NIMBUS is useful when the DM can express directional feedback about objectives more naturally than exact trade-offs.

Navigation methods emphasize learning by moving through the nondominated set rather than by repeatedly asking for complete preference statements. NAUTILUS starts from an easily understood inferior point and iteratively moves toward the nondominated set, allowing the DM to guide the search while seeing how objective values improve and what trade-offs become unavoidable \citep{miettinen2010nautilus}. As a representative approach based on pairwise comparisons, the interactive approximation algorithm of \citet{lokman2018interactive} elicits the DM's preferences through comparisons of pairs of solutions and uses the resulting preference cones to reduce the search space for multi-objective integer programs.

Overall, the literature shows that interactive methods can reduce the need to present a large nondominated set, but they also assume that the DM can provide stable and method-specific preference information. For expert users this may be appropriate. For business stakeholders, the interaction may still feel technical: they may prefer to ask qualitative questions, request explanations, and refine preferences conversationally.

\subsection{LLMs in optimization}
Recent work assigns LLMs several distinct roles in decision support systems: explaining solver outputs, translating requests into model changes, eliciting preferences, and guiding the search itself. These roles address different parts of the decision-support process and provide complementary foundations for conversational optimization. For example, OptiGuide, proposed by \citet{li2023supplychain}, connects natural-language what-if questions to executable code and an optimization solver, then interprets the resulting outputs for the user. Its emphasis is on obtaining grounded insights from an existing supply-chain model, illustrating the value of combining language generation with numerical computation performed by dedicated tools. \citet{ye2026patches} translate user requests into structured model patches and select re-optimization techniques that exploit previous solutions and solver information, with applications to supply-chain recovery and university exam scheduling. Their central concern is efficient adaptation of an existing model to changed requirements. In contrast, \citet{ge2026clarification} address missing information before formulation: their OR-Clarify benchmark and InterOPT framework evaluate which questions an agent should ask, when it should stop asking, and whether it silently assumes unspecified business rules. This distinction matters because clarifying the definition of a problem and eliciting trade-offs within a defined problem are related but different tasks.

A related stream places preference elicitation within an iterative optimization loop. \citet{lawless2024interactive} combine LLMs with constraint programming for meeting scheduling, allowing users to articulate contextual preferences and revise them after inspecting proposed solutions. \citet{ghiani2024semistructured} propose a methodology for semi-structured decision making in which feedback can modify objective functions, constraints, or heuristic parameters, and illustrate it through last-mile delivery system design. LAPPI, introduced by \citet{kuroki2025lappi}, focuses on an earlier stage: conversationally enumerating candidate items and assigning preference scores and numerical attributes to instantiate a combinatorial optimization problem. These studies establish that natural-language interaction, iterative preference refinement, and the separation of language processing from optimization are already important features of existing systems. \citet{ghiani2025adaptability} combine dynamically generated evaluation metrics with an LLM-supported MCDM module that selects among alternative plans, demonstrating the approach on a rich vehicle-routing problem. 

Natural-language preference elicitation has also been studied as a selection problem over available alternatives. PEBOL, proposed by \citet{austin2024bayesian}, maintains Bayesian beliefs about item preferences and uses acquisition strategies to guide LLM-generated questions. At each turn, the user answers the generated query in free-form natural language, and natural-language inference is used to update the Bayesian preference beliefs. LISTEN, developed by \citet{jovine2026listen}, instead uses an LLM prompted with natural-language preference description to simulate the DM's preferences to select from a large set of multi-attribute alternatives. LISTEN-U iteratively refines a linear utility representation, whereas LISTEN-T selects candidates through a tournament without imposing that parametric form. In its exam-scheduling case study, LISTEN follows an \emph{a posteriori} approach by selecting from a large set of schedules generated before preference-based selection begins. ParetoChat instead uses the DM's evolving preferences to generate targeted candidates through successive scalarizations, avoiding the need to enumerate the full Pareto set in advance, which can be computationally demanding and increasingly difficult as the number of objectives grows. 

Language-based feedback can also guide optimization beyond a fixed candidate set. LILO, introduced by \citet{kobalczyk2026lilo}, studies a preference-based Bayesian optimization setting. LILO converts free-form feedback and prior knowledge into pairwise preference signals, fits a Gaussian process preference model, and uses Bayesian optimization acquisition functions to choose subsequent experiments. It therefore shares the principle of using an LLM as a preference interface while assigning search to a separate optimization procedure. MOHOLLM, proposed by \citet{schwanke2026multiobjective}, assigns the LLM a different role: candidate generation and surrogate prediction within an adaptive hierarchical search for a Pareto front. Its focus is multi-objective black-box search performance rather than conversational elicitation of a DM's preferred compromise. 

Evaluating complete conversations is another directly relevant research direction. \citet{drossman2026conversation} develop LLM-based decision agents with internal utility functions to evaluate optimization agents through multi-turn interactions. Their school-scheduling study compares general-purpose and increasingly structured optimization agents and shows the importance of assessing the full conversation rather than a single response. This work is particularly close to our simulation methodology: both use artificial DMs and utility-based assessment to study conversational optimization. 

Against this background, ParetoChat's contribution lies in the combination of a scalarization-based conversational interface and a comparative evaluation within interactive multi-objective optimization. Given a predefined feasible set and objective functions, ParetoChat translates qualitative requests and references to earlier solutions into parameters of supported scalarization methods, then invokes the existing solver to generate further candidates. The emphasis is on navigating trade-offs within that model, rather than instantiating the underlying problem, rewriting its operational rules, selecting only from a precomputed candidate set, or learning a Gaussian process preference surrogate. 

\section{ParetoChat: An LLM-Based Expert Agent}
\label{sec:paretochat}

\subsection{Problem definition}
\label{sec:probdef}
We consider the problem of translating a DM's request for a new solution, expressed in natural language, into a scalarization function. Let $F$ be the finite set of scalarization functions supported by the optimization solver. Each $f \in F$ maps an objective vector in $\mathbb{R}^m$ to a scalar in $\mathbb{R}$. Given the MOOP in Equation~\ref{eq:moop}, each scalarization defines a single-objective problem that the solver attempts to solve:

\begin{equation}
\label{eq:singleobjprob}
P(f): \max_{x \in X} \; f(z(x))
\end{equation}

Let $x^*(f)$ denote a solution returned by the solver for $P(f)$, and let $z^*(f)=z(x^*(f))$ be its objective vector. The DM assigns the value $V(z^*(f))$ to this outcome and seeks to maximize it by choosing among the available scalarizations:

\begin{equation}
\label{eq:dmvaluemax}
P^{DM}: \max_{f \in F} V(z^*(f))
\end{equation}

Because every returned solution is feasible, $\max_{f \in F} V(z^*(f)) \leq \max_{x \in X} V(z(x))$. Equality holds when the available scalarizations and solver can produce at least one objective vector that maximizes $V(z(x))$ over $x \in X$. Thus, the outcomes available to the DM depend both on the scalarizations in $F$ and on the solver's ability to optimize each corresponding problem. For example, weighted-sum scalarizations generally recover only supported nondominated points, whereas suitable achievement scalarizing functions can represent the full nondominated set, assuming exact optimization.

Because DMs may find it difficult to specify a scalarization function that reflects their preferences directly, we seek a mapping $L$ from a natural-language request $u$ to a supported scalarization, $f=L(u) \in F$. Let $\mathcal{U}$ denote the set of possible requests. Ideally, $L$ enables the DM to attain the best value achievable through any scalarization in $F$:

\begin{equation}
\label{eq:llmmapping}
\max_{u \in \mathcal{U}} V(z^*(L(u))) = \max_{f \in F} V(z^*(f))
\end{equation}

\subsection{Method}
\label{sec:method}
In our approach, we use an LLM-based agent as a mapping function $L$. Algorithm \ref{alg:paretochat} presents how our method interacts with the user.

\begin{algorithm}[H]
\caption{ParetoChat interaction loop}
\label{alg:paretochat}
\KwIn{Problem information $I$ and an optimization solver}
\KwOut{Solution pool $S$ and the corresponding solution history $H$}
Initialize $H_0 = \emptyset$, $S_0 = \emptyset$, and $t = 0$\;
\While{the DM has not terminated the session}{
  $t = t + 1$\;
  Receive a new utterance $u_t$\;
  \uIf{$u_t$ requests a new solution}{
    Compute $f_t = L(u_t \mid I, H_{t-1}, S_{t-1})$\;
    Solve $P(f_t)$ using the optimization solver\;
    \eIf{a feasible solution is found}{
      Set $s_t$ to the solution and explain its objective values and how $u_t$ was translated into $f_t$\;
      Update $S_t = S_{t-1} \cup \left\{s_t\right\}$
    }{
      Set $s_t = \mathrm{null}$ and explain why no solution was found\;
    }
    Let $h_t=\left\{u_t, f_t, s_t\right\}$\;
    Update $H_t = H_{t-1} \cup \left\{h_t\right\}$.
    
  }
  \uElseIf{$u_t$ provides feedback on a previous solution $s_{t'}$}{
    Add $u_t$ as a feedback to the context of this solution, $h_{t'}=h_{t'} \cup \left\{u_t\right\}$ \;
  }
  \uElseIf{$u_t$ requests a session summary}{
    Generate and return a summary using $H_{t-1}$ and $S_{t-1}$\;
  }
  \Else{
    Mark the session as terminated\;
  }
}
\end{algorithm}

\subsection{System overview}
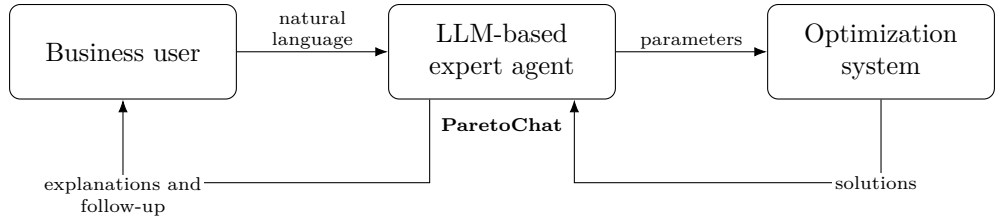
\begin{figure}[htbp]
\centering
\begin{tikzpicture}[
  node distance=2.0cm,
  box/.style={draw, rounded corners, align=center, minimum width=3.0cm, minimum height=1.25cm},
  lab/.style={font=\scriptsize, align=center, fill=white, inner sep=1pt}
]
\node[box] (user) {Business user};
\node[box, right=of user] (agent) {LLM-based\\expert agent};
\node[below=0.12cm of agent, font=\bfseries\scriptsize] {ParetoChat};
\node[box, right=of agent] (solver) {Optimization\\system};

\draw[-{Latex}] (user) -- node[lab, above]{natural\\language} (agent);
\draw[-{Latex}] (agent) -- node[lab, above]{parameters} (solver);
\draw[-{Latex}] (solver.south) |- ++(0,-1.1) -| node[lab, pos=0.08, right]{solutions} ([xshift=-0.55cm]agent.south east);
\draw[-{Latex}] ([xshift=0.55cm]agent.south west) |- ++(0,-1.1) -| node[lab, pos=0.55, below]{explanations and\\follow-up} (user.south);
\end{tikzpicture}
\caption{ParetoChat acts as a natural-language interaction layer between business users and optimization systems.}
\label{fig:overview}
\end{figure}

ParetoChat is designed as an interaction layer between a business user and an optimization system. Figure~\ref{fig:overview} illustrates the basic loop. The user provides natural-language requests, the agent translates those requests into solver-compatible parameters, the optimization system then runs with those parameters and returns one or more candidate solutions, and the agent explains the result and asks for feedback. The process continues as the user reviews the generated solutions, gives feedback on them, and asks for further solutions or a summary of the solutions generated so far.

We assume that the optimization system has built-in objective scalarization methods with configurable parameters (see Appendix~\ref{app:scalarization-methods}). The agent is then given information about how it can run the optimizer by selecting a scalarization method and configuring its parameters. ParetoChat limits itself to the objective space and focuses on transforming DMs' preferences regarding the objectives into available scalarization parameters. It does not allow users to arbitrarily update the problem definition or update the internals of the optimization solver (e.g. adding new constraints, changing a heuristic). This well-scoped design allows ParetoChat to support different multi-objective problems and optimization solvers (e.g. math programs,  heuristics) with minimal contextual info, provided that the solver exposes configurable scalarization methods and returns the resulting objective values of the generated solutions. The optimization problem metadata is supplied in a structured form, including a list of objectives with their names, descriptions, and optimization senses (minimization or maximization), and which scalarization types the optimization solver supports.

\subsection{Multi-agent architecture}
The system follows a multi-agent architecture implemented with the LangGraph agent development framework \citep{langgraph2024}. The multi-agent design makes the system easier to test and debug. It allows separation of concerns in the overall workflow so that specific parts of the flow can be improved independently.

\begin{figure}[ht]
  \centering
  \resizebox{\textwidth}{!}{%
  \begin{tikzpicture}[
    node distance=0.62cm and 0.45cm,
    box/.style={draw, rounded corners, align=center, minimum width=2.2cm, minimum height=0.7cm, font=\tiny},
    ctx/.style={draw, rounded corners, align=center, minimum width=2.55cm, minimum height=0.7cm, font=\tiny, fill=gray!8},
    database/.style={
      cylinder,
      shape border rotate=90,
      aspect=0.28,
      draw,
      align=center,
      minimum width=1.55cm,
      minimum height=1.5cm,
      font=\tiny,
      fill=gray!8
    },
    person/.style={
      draw,
      circle,
      minimum size=0.95cm,
      inner sep=0pt,
    }
  ]
  \node[person] (user) {DM};
  \node[box, right=1.2cm of user] (intent) {User Intent Agent};
  \node[box, below=of intent] (command) {Command Processor Agent};
  \node[box, right=0.5cm of command] (feedback) {Feedback Processor Agent};
  \node[box, below=of command] (expert) {MCDM Expert Agent};
  \node[box, left=0.5cm of command] (summary) {Session Summarizer Agent};

  \node[ctx, left=1.0cm of expert] (problem) {Problem Info};
  \node[ctx, right=1.5cm of expert] (solver) {Optimization Solver};
  \node[database, below=of expert] (solpool) {Solution\\ Pool};

  \node[box, left=0.5cm of solpool] (present) {Solution Presenter Agent};

  \draw[-{Latex}] (user.east) -- node[above, font=\tiny] {request} (intent.west);
  \draw[-{Latex}] (intent.south) -- node[right, font=\tiny] {command} (command.north);
  \draw[-{Latex}] (intent.east) -- node[above, font=\tiny] {feedback} ++(2.4,0) -| (feedback.north);
  \draw[-{Latex}] (command.south) -- (expert.north);
  \draw[-{Latex}] (command.west) -- (summary.east);
  \draw[-{Latex}] (feedback.east) -- ++(0.5,0) |- (solpool.east);
  \draw[-{Latex}] (solpool.north) -- (expert.south);
  \draw[-{Latex}] (problem.east) -- (expert.west);
  \draw[-{Latex}] (expert.east) -- node[above, font=\tiny] {scalarization} (solver.west);
  \draw[-{Latex}] (solver.south)
    -- node[midway, left, font=\tiny] {solution}
    (solver.south |- solpool.east)
    -- (solpool.east);
  \draw[-{Latex}] (summary.north)
    -- node[midway, left, font=\tiny] {summary}
    (summary.north |- user.west)
    -- (user.west);
  \coordinate (presentreturn) at ([xshift=-1.0cm]present.west);
  \draw[-{Latex}] (present.west)
    -- (presentreturn)
    -- node[midway, above, rotate=90, font=\tiny] {explanation}
    (presentreturn |- user.west)
    -- (user.west);
  \draw[-{Latex}] (solpool.west) -- (present.east);
  \end{tikzpicture}%
  }
  \caption{ParetoChat's multi-agent architecture}
  \label{fig:multi-agent-arch}
  
\end{figure}
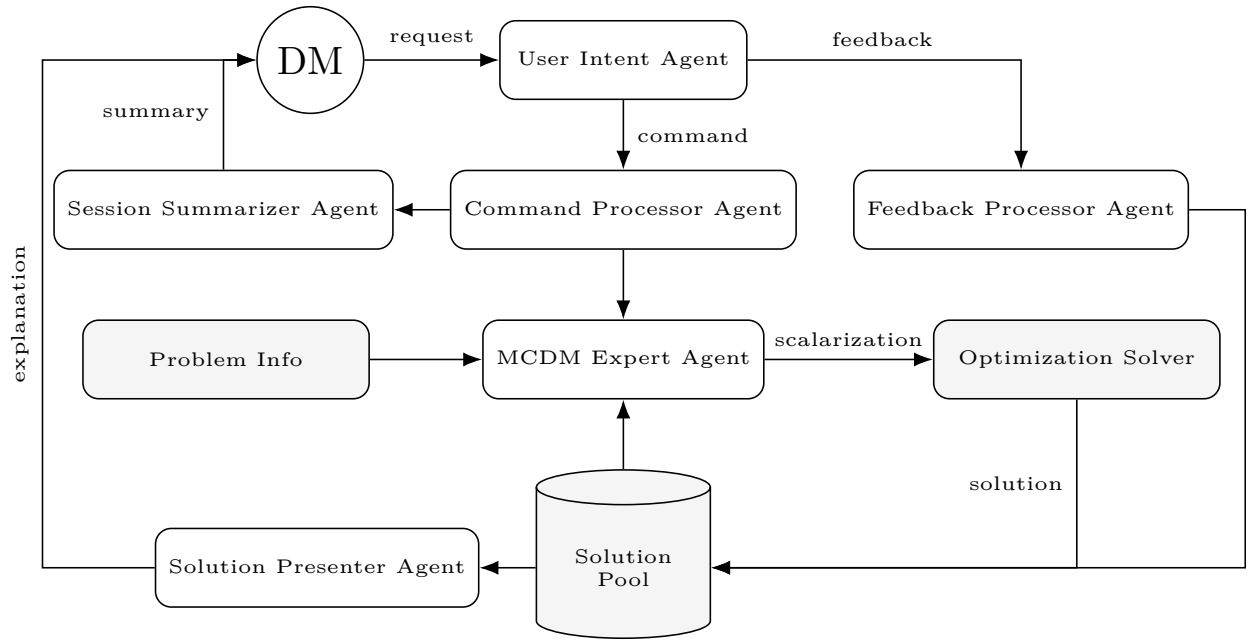

Figure~\ref{fig:multi-agent-arch} summarizes the main sub-agents (nodes) and information flows (edges) in the LangGraph graph implementation. The implementation maintains the session state (e.g., message history, tool calls, and previous solution iterations) to provide context for the next user input. Some edges are conditional: the next agent depends on the output of the previous agent (e.g., the user intent agent may route processing to either the command processor or the feedback processor). The role of each agent is described below. The prompts used for each agent are given in Appendix~\ref{app:agent-prompts}.

\subsubsection{User intent agent}
The User Intent Agent is the first component that processes each user message. Its task is to identify the communicative intent of the input before any optimization tool is called. In particular, it distinguishes between requests for a new solution, preference updates, feedback about a displayed solution, requests to save or reject a solution, and requests for a session summary. This classification reduces ambiguity for downstream agents: a sentence such as ``this is too expensive, try something faster'' is interpreted both as feedback on the current solution and as a preference update toward lower delivery time. The output of this agent is a structured output object that includes the original user input and the detected intent types.

\subsubsection{Command processor agent}
Given the structured intent produced by the User Intent Agent, the Command Processor Agent determines which action should be executed next based on the nature of the user command. For example, it routes ``find a new solution'' requests to the MCDM Expert Agent and sends ``summarize our progress'' requests to the Session Summarizer Agent.

\subsubsection{MCDM expert agent}
This agent contains the MCDM expertise of the system. It receives the user's expressed preferences, the problem metadata, and the list of available tools, and converts them into supported scalarization parameters. Depending on the user request, this may involve selecting a weighted-sum method, constructing a reference point, setting lexicographic priorities among the objectives, or setting upper or lower bounds on the objectives. Allowing users to set upper (lower) bounds on the objectives for maximization (minimization) problems may lead to generating dominated points. However, this functionality allows users to express requests such as finding a new solution between the last two solutions, helping them concentrate their search on specific parts of the objective space. Users can always ask whether another solution dominates the current solution to check whether it can be improved in at least one objective without worsening any other.

Several tools are made available to this agent, and the prompt instructs it when to use each of them. By calling them as needed, the agent retrieves the problem context and the current state of the solution pool. Then, it returns the inferred scalarization as a structured output, including the scalarization type (e.g., weighted-sum, lexicographic, or achievement scalarizing function), scalarization parameters (e.g., weights, priorities, or a reference point), and bounds for each objective, if there exist any.

In Table~\ref{tab:mcdm-agent-scalarization-examples}, we present some example user preference inputs and the corresponding scalarizations we might expect the agent to return for a bi-objective discrete-choice problem where the first and second objectives are cost and delivery time minimization, respectively. See Appendix \ref{app:scalarization-methods} for the definition of scalarizations and their parameters. Note that the MCDM Expert Agent is instructed to limit its output only to the scalarization types that are supported by the optimization solver.

\begin{table}[ht]
\caption{Examples of user preferences and likely structured scalarization outputs from the MCDM expert agent for a bi-objective minimization problem.}
\label{tab:mcdm-agent-scalarization-examples}
\centering
\scriptsize
\renewcommand{\arraystretch}{1.15}
\begin{tabular*}{\textwidth}{@{\extracolsep{\fill}}p{0.30\textwidth}p{0.17\textwidth}p{0.18\textwidth}p{0.24\textwidth}@{}}
\toprule
\textbf{User preference} & \textbf{Referenced solutions} & \textbf{Scalarization type} & \textbf{Scalarization parameters} \\
\midrule
Find a balanced solution across both objectives. & None & Weighted-sum & $w=(0.5, 0.5)$ \\
Improve delivery time, but do not let cost increase too much. & $z = (1000, 60)$ & Augmented epsilon constraint & $q=2$, $\epsilon=1100$, $\rho=10^{-4}$ \\
Cost is most important, then delivery time. & None & Lexicographic & $priority_{cost}$=1, $priority_{delivery\_time}$=2 \\
Show me something between these two saved options. & $z^1=(1000, 60)$, $z^2=(1200, 40)$ & Weighted-sum in a bounded region & $w=(0.5, 0.5)$, $1000 \leq z_1 \leq 1200$, $40 \leq z_2 \leq 60$ \\
Try to reach 30 minutes delivery time while keeping other metrics reasonable. & None & Achievement scalarizing function & $z^r = (1200, 30)$, $w=(0.5, 0.5)$, $\rho=10^{-4}$ \\
\bottomrule
\end{tabular*}
\end{table}

\subsubsection{Optimization solver}
The optimization solver is not an AI agent but a mathematical program or a heuristic algorithm that ParetoChat integrates with to generate a solution for the single-objective version of the problem. We assume that such a solver exists and is capable of working with the scalarization methods provided by the MCDM Expert Agent to generate a corresponding solution. We also assume that objectives are properly scaled on the solver side, and we do not give the MCDM Expert Agent any prior information about their scales.

In our graph implementation, the optimization solver is triggered with the scalarization output of the MCDM Expert Agent and is expected to return a solution. If there is a feasible solution, the returned solution is then compared against the existing solution pool to check whether it has already been generated in a previous iteration. If a new solution could not be found, the MCDM Expert Agent runs again to generate a perturbed scalarization. We try a maximum of three times to generate a different solution for the user. Note that we do not check whether the new solution is dominated by any of the existing solutions in the solution pool.

\subsubsection{Solution presenter agent}
The Solution Presenter Agent converts the raw output of the optimization solver into natural language to support the DM's decision making. It receives the solution, the scalarization settings used to generate it, and relevant solver status information. If a solution is returned, it produces a concise explanation clarifying why this particular alternative was generated. It also addresses cases where the solver could not find a solution.

\subsubsection{Feedback processor agent}
The Feedback Processor Agent interprets the user's reaction after a candidate solution has been presented. It detects whether the user accepts the solution, rejects it, wants to save it for later comparison, or expresses a new direction for the search. It can also identify frustration or uncertainty signals, such as repeated rejections or vague statements like ``this is still not balanced enough.'' The processed feedback is converted into structured output including the feedback sentiment and whether the solution should be saved in the solution pool for later review. The collected feedback is used as additional context for the MCDM Expert Agent in the generation of new scalarizations. This allows the system to support iterative preference elicitation: each reaction becomes additional evidence about the user's preferred region of the nondominated set.

\subsubsection{Session summarizer agent}
This agent maintains a compact representation of the interaction history. It records the solutions shown to the user, saved alternatives, rejected alternatives, scalarization choices, and the evolution of expressed preferences. When requested, it can produce a user-facing summary of the current session, including what has been tried and why certain alternatives were preferred or discarded.

\subsection{Example interaction with ParetoChat}
\begin{figure}[ht]
\centering
\begin{tikzpicture}
\begin{axis}[
  width=0.88\textwidth,
  height=0.48\textwidth,
  xlabel={Cost ($\times 10^3$ Euros)},
  ylabel={Delivery time (mins)},
  xmin=0,
  xmax=105,
  ymin=28,
  ymax=52,
  grid=major,
  axis lines=left,
  tick align=outside,
  scaled x ticks=false,
  xticklabel style={/pgf/number format/fixed,
                    /pgf/number format/1000 sep={,}},
]
\addplot[
  only marks,
  mark=*,
  mark size=2.0pt,
  blue!65!black,
] coordinates {
  (101.0,30.0)
  (100.0,30.0)
  (25.0,34.0)
  (25.0,34.8)
  (15.0,35.0)
  (14.0,35.0)
  (9.0,36.0)
  (8.0,37.0)
  (6.0,38.0)
  (5.5,39.0)
  (5.4,40.0)
  (5.2,45.0)
  (5.0,50.0)
  (5.0,49.0)
};
\addplot[
  only marks,
  mark=square,
  mark options={line width=1.2pt},
  mark size=2.5pt,
  red,
] coordinates {
  (9.0,36.0)
  (25.0,34.0)
  (14.0,35.0)
};
\node[red, font=\bfseries\scriptsize, anchor=south west,
      xshift=2pt, yshift=2pt] at (axis cs:9.0,36.0) {1};
\node[red, font=\bfseries\scriptsize, anchor=south west,
      xshift=2pt, yshift=2pt] at (axis cs:25.0,34.0) {2};
\node[red, font=\bfseries\scriptsize, anchor=south west,
      xshift=2pt, yshift=2pt] at (axis cs:14.0,35.0) {3};
\end{axis}
\end{tikzpicture}
\caption{Feasible points for the example problem. The points presented by ParetoChat during the user session are marked with hollow red squares together with their display sequence.}
\label{fig:example-problem-solutions}
\end{figure}
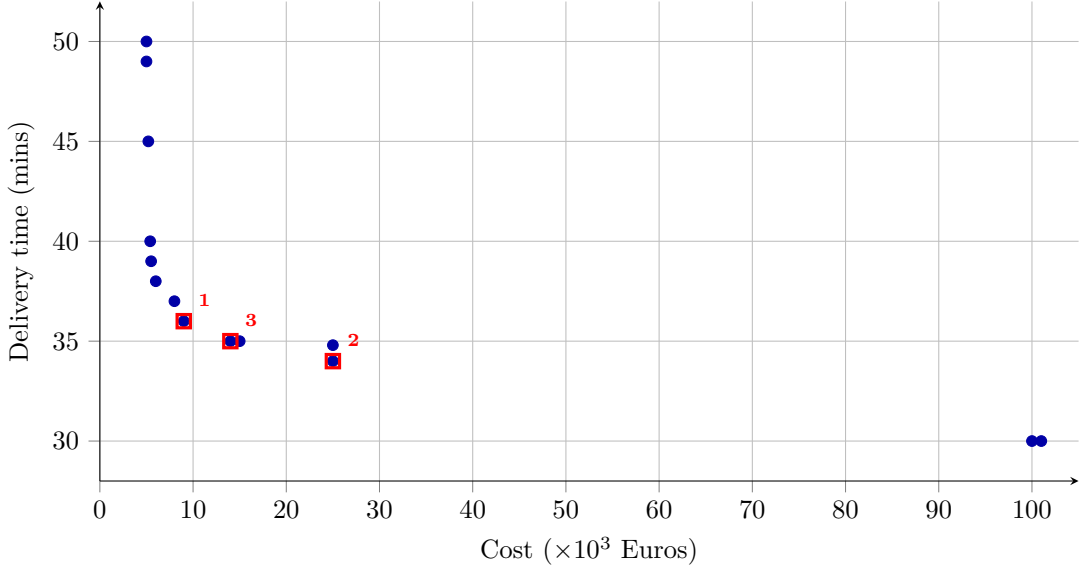

We consider a bi-objective problem with cost and delivery time as objectives, both to be minimized. The feasible solutions and the corresponding objective values are precomputed for the sake of this example, and the task is to find the user's most-preferred solution among them. Table~\ref{tab:example-problem-solutions} lists the feasible set of points, some of which are dominated. Figure~\ref{fig:example_problem_info} shows the ProblemInfo configuration supplied to ParetoChat. The optimization solver in this case is a simple routine: it applies min--max scaling to the available points to bring both objectives onto a common scale, then selects the point that optimizes the given scalarization function. The points are visualized in Figure~\ref{fig:example-problem-solutions}. Note that the user does not know the precomputed solutions beforehand.

\begin{table}[ht]
\centering
\caption{Feasible points for the example problem}
\label{tab:example-problem-solutions}
\begin{tabular}{rr}
\toprule
\textbf{Cost (Euros)} & \textbf{Delivery time (mins)} \\
\midrule
101000.0 & 30.0 \\
100000.0 & 30.0 \\
25000.0 & 34.0 \\
25000.0 & 34.8 \\
15000.0 & 35.0 \\
14000.0 & 35.0 \\
9000.0 & 36.0 \\
8000.0 & 37.0 \\
6000.0 & 38.0 \\
5000.5 & 39.0 \\
5000.4 & 40.0 \\
5000.2 & 45.0 \\
5000.0 & 50.0 \\
5000.0 & 49.0 \\
\bottomrule
\end{tabular}
\end{table}

\begin{figure}[htbp]
\centering
\begin{Verbatim}[
  fontsize=\footnotesize,
  breaklines=true,
  breakanywhere=true,
  breaksymbolleft={},
  breaksymbolright={},
  frame=single,
  framerule=0.4pt,
  framesep=2mm,
  rulecolor=\color{black!55}
]
problem:
  description: """
    This is a bi-objective discrete-choice problem. The task is
    to select the best solution from a set of solutions based on
    the user's preferences.
  """
  objectives:
    - name: "cost"
      description: "Total cost of delivering all current orders"
      sense: "MINIMIZE"
    - name: "delivery_time"
      description: "Average delivery time per order"
      sense: "MINIMIZE"
algorithm:
  name: "Discrete Choice"
  description: "Select the best of the available solutions"
  optimization_direction: "MINIMIZE"
  supports_obj_bounds: true
  supported_scalarizations:
    - "WEIGHTED_SUM"
    - "LEXICOGRAPHIC"
    - "REFERENCE_POINT"
\end{Verbatim}
\caption{ProblemInfo configuration for the example interaction}
\label{fig:example_problem_info}
\end{figure}

\FloatBarrier

Table~\ref{tab:example-illustration} presents the four-round interaction between the user and ParetoChat. The user starts by searching for a solution that performs well on both objectives, balancing them to some extent. The agent translates this request into a weighted-sum scalarization with equal weights, and the optimizer finds Solution 1 in Figure~\ref{fig:example-problem-solutions}. This solution lies around the middle of the spectrum and is a good starting point for a user who wants to avoid extreme solutions. The user is then interested in seeing the potential trade-offs if the delivery time could be reduced further to around 30 mins. The agent uses this value as a reference point and finds Solution 2, with an improved delivery time compared to Solution 1. Note that the agent does not directly generate the extreme nondominated point $(10^5, 30)$, because the user is more interested in balanced solutions, as expressed in the first iteration.

Finding the cost increase from Solution 1 to Solution 2 too large, the user then looks for a better trade-off in the region around Solution 1. The agent interprets this \say{in-between} query by constructing bounds corresponding to the objective values of those two solutions and restricting the new search to this region. The optimizer then finds Solution 3 in this region. The user seems more satisfied with the trade-off between Solutions 1 and 3 than with that between Solutions 1 and 2, and wants to save Solution 3 and review the progress so far.

It is important to note that the agent successfully avoids presenting a dominated point (right next to Solution 3) to the user in this case. However, that is not guaranteed in all cases. Since we allow the MCDM Expert Agent to use both upper and lower bounds in its scalarizations, the generated points might be dominated, although they are nondominated in the bounded objective space.

\newcommand{\usercard}[1]{%
  \fcolorbox{blue!50!black}{blue!6}{%
    \parbox[t][1.15cm][t]{0.42\textwidth}{\raggedright
    {\color{blue!50!black}\textbf{User}}\\[0.15em]#1}}
}
\newcommand{\agentcard}[2]{%
  \fcolorbox{green!45!black}{green!6}{%
    \parbox[t][#1][t]{0.42\textwidth}{\raggedright
    {\color{green!35!black}\textbf{ParetoChat}}\\[0.15em]#2}}
}

\begin{table}[ht]
\centering
\scriptsize
\setlength{\fboxsep}{4pt}
\begin{tabular}{@{}>{\centering\arraybackslash}m{0.445\textwidth}@{\hspace{0.001\textwidth}}>{\centering\arraybackslash}m{0.445\textwidth}@{}}
{\color{gray!70}\textbf{Round 1}} & {\color{gray!70}\textbf{Round 2}} \\
[0.3em]
\usercard{Find a solution that balances both objectives equally.}
&
\usercard{Nice save it and find a solution that improves the delivery time somewhere around 30 minutes.}
\\[0.25em]
\agentcard{3.95cm}{Based on your request to find a solution that balances both objectives equally, I found a candidate solution for you.\\[0.25em]
To achieve this balance, I used the \textbf{weighted-sum} method. This assigns equal importance, or a weight of 50\%, to both the cost and delivery-time objectives.\\[0.25em]
The resulting solution has the following objective values.\\[0.25em]
\textbf{Cost:} 9,000.0\\
\textbf{Delivery Time:} 36.0\\[0.25em]
What do you think of this solution? Would you like to save it?}
&
\agentcard{3.95cm}{Of course! Here is a presentation of the solution found based on your latest request.\\[0.25em]
Based on your request to find a solution with a delivery time around 30 minutes, I have found a new solution for you.\\[0.25em]
To achieve this, I used a \textbf{reference point}-based method. This means I searched for a solution where the delivery time is as close as possible to your target of 30 minutes.\\[0.25em]
Here is the solution I found:\\[0.25em]
\textbf{Cost:} 25,000.0\\
\textbf{Delivery Time:} 34.0}
\end{tabular}

\scriptsize
\setlength{\fboxsep}{4pt}
\begin{tabular}{@{}>{\centering\arraybackslash}m{0.445\textwidth}@{\hspace{0.001\textwidth}}>{\centering\arraybackslash}m{0.445\textwidth}@{}}
{\color{gray!70}\textbf{Round 3}} & {\color{gray!70}\textbf{Round 4}} \\
[0.3em]
\usercard{The cost increased too much. Is there any other solution in between?}
&
\usercard{Ok, let's save this as well. Now, summarize our progress so far.}
\\[0.25em]
\agentcard{4.25cm}{Based on your request for an intermediate solution, I have found a new option for you.\\[0.25em]
To find this, I used a \textbf{weighted-sum} approach where I assigned equal importance to both cost and delivery time. I also focused the search on solutions with a cost between approximately 9,000 and 25,000, and delivery time between 34 and 36 mins.\\[0.25em]
The resulting solution is:\\[0.25em]
\textbf{Cost:} 14,000.0\\
\textbf{Delivery Time:} 35.0\\[0.25em]
What do you think of this solution? Do you have any feedback, and would you like to save it?}
&
\agentcard{4.25cm}{Of course! Here is a summary of our progress so far.\\[0.25em]
\textbf{Session Summary:}\\
So far, you have explored 3 solutions and saved 2 of them. Here's a quick rundown of what we've covered:\\[0.25em]
\textbf{Key Decisions and Trade-offs:}\\
$\bullet$ We started by finding a balanced solution, which you saved.\\
$\bullet$ You then tried to improve the delivery time, but found the increase in cost to be too high.\\
$\bullet$ Finally, we found a middle-ground solution that you also saved.\\[0.25em]
This has helped us understand the trade-off between delivery time and cost. \textit{[continues \ldots]}}
\end{tabular}
\caption{A real conversation with ParetoChat on the example problem.}
\label{tab:example-illustration}
\end{table}

\normalsize

\section{Simulation Study}
\label{sec:simulation}
In this section, we first describe the framework used to evaluate ParetoChat against some established interactive multi-objective optimization methods. The framework is designed to support controlled and reproducible experiments while capturing both the quality of the final preferred solution and the interaction effort required to reach it. 

\begin{figure}[h]
\centering
\begin{tikzpicture}[
  font=\small,
  node distance=0.55cm and 0.7cm,
  >=Latex,
  box/.style={draw, rounded corners, align=center, fill=blue!4, minimum height=0.8cm, text width=3.2cm},
  aidmbox/.style={draw, rounded corners, align=center, fill=red!18, draw=red!65!black, line width=0.9pt, minimum height=0.85cm, text width=3.5cm},
  wide/.style={draw, rounded corners, align=center, fill=green!5, minimum height=0.85cm, text width=7.0cm},
  tool/.style={draw, rounded corners, align=left, fill=orange!6, text width=4.2cm, inner sep=3pt},
  metric/.style={draw, rounded corners, align=left, fill=purple!6, text width=4.0cm, inner sep=3pt},
  note/.style={font=\scriptsize, align=center}
]

\node[box] (repo) {\textbf{Problem repo}};
\node[wide, right=of repo] (prompt) {\textbf{Prompt}\\problem info + solution-tool user guides\\\textbf{Instruction:} first explore, then exploit regions of interest};
\node[aidmbox, below=of prompt] (aidm) {\textbf{AIDM}\\Acts as a human DM\\searching for a preferred solution};

\node[tool, below=1.0cm of aidm, xshift=-4.8cm] (solution) {\textbf{Solution tools}\\Find next solution(s) for the DM's request\\-- ParetoChat\\-- DESDEO RPM\\-- DESDEO NIMBUS};
\node[tool, below=1.0cm of aidm] (eval) {\textbf{Evaluation tool}\\Calculate the value of the currently considered solution\\-- linear value function\\-- nonlinear value function};
\node[tool, below=1.0cm of aidm, xshift=4.8cm] (update) {\textbf{Decision update tool}\\Update the current most preferred solution};

\node[metric, below=0.35cm of eval] (perf) {\textbf{Performance evaluation}\\-- value gap\\-- run time\\-- number of evaluations};

\draw[->] (repo) |- node[note, above right]{instance} (aidm);
\draw[->] (prompt) -- (aidm);
\draw[->] (aidm) -- (solution);
\draw[->] (solution) -- node[note, below left, xshift=-1cm]{candidate solutions} (aidm);
\draw[->] (aidm) -- (eval);
\draw[->] (eval) -- node[note, right]{solution value} (aidm);
\draw[->] (aidm) -- (update);
\draw[->] (update) -- node[note, below right, xshift=1cm]{preferred solution updated} (aidm);
\draw[->] (solution.south) |- (perf.west);
\draw[->] (eval.south) -- (perf.north);
\draw[->] (update.south) |- (perf.east);
\end{tikzpicture}
\caption{The architecture of our evaluation framework based on an AI-based DM}
\label{fig:AIDM}
\end{figure}
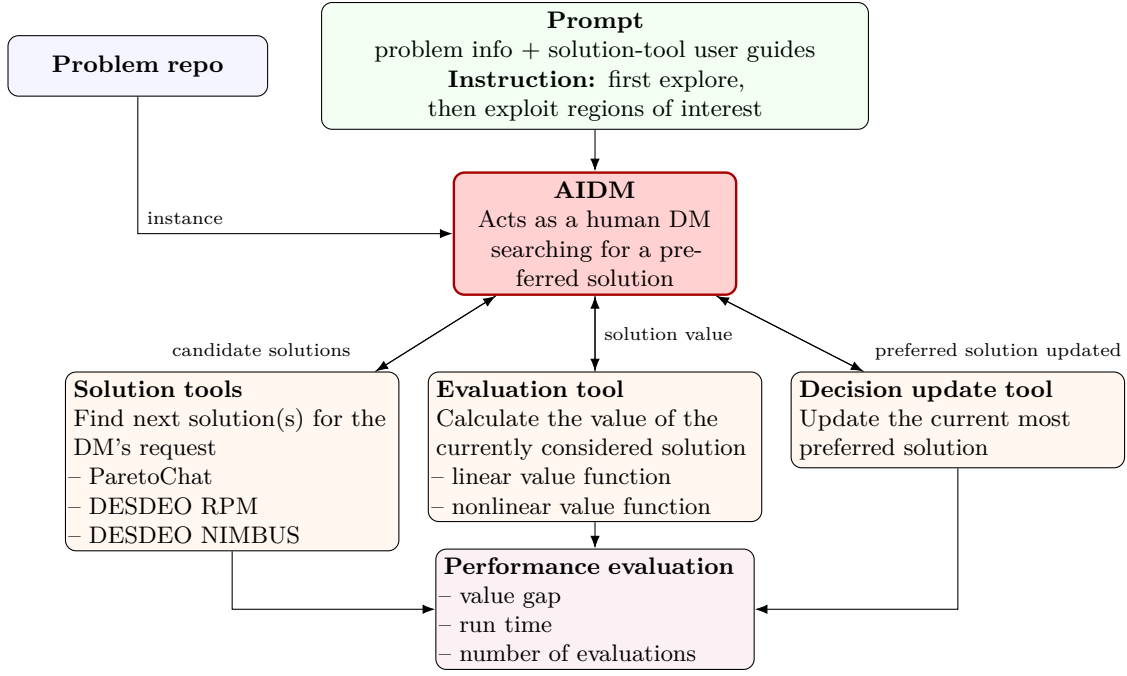

In Figure \ref{fig:AIDM}, we present the main components in the system. It consists of a problem repository, an AI-based agent to act as a human DM, solution tools available for benchmarking, an evaluation tool that computes hidden utility values for candidate solutions, a decision update tool to be used by the agent to keep track of its most preferred solution so far and a performance evaluation module to calculate and report the performance measures for an experiment run.

We first introduce the artificial decision maker, then describe the benchmark problems and compared methods, and finally report the experiment results.

\subsection{AI agent as a decision maker}
Evaluating interactive multi-objective optimization methods with human subjects is expensive and difficult to reproduce. Following the spirit of prior experimental designs for interactive methods \citep{afsar2024experimental} and artificial decision maker (ADM) studies \citep{afsar2023comparing}, we use an ADM to compare methods under controlled conditions. 

Unlike previous studies, we use an AI agent-based DM (AIDM), rather than a static, predetermined value function alone, to represent a DM's behavior. The AIDM is given problem information and user guides for the compared methods (solution tools). It interacts with a method iteratively, requesting candidate solutions, evaluating them according to a hidden value function, and updating its currently preferred solution.

It is important to note that the value function is exposed neither to the solution method nor to the AIDM. That is, the AIDM does not know a priori what value function it needs to optimize to find its most-preferred solution. It can only associate a value with a presented solution using the solution evaluation tool and can therefore develop an implicit understanding of its value function as it explores different solutions. The AIDM is instructed to first explore and then exploit regions of interest, mimicking a human DM who gradually learns what trade-offs are available.

The prompt used for the AIDM is given in Appendix~\ref{app:dm-agent-prompt}. The prompt includes placeholders for information about the objectives, the tool the agent needs to use to generate a solution, and its user guide. The user guide for a solution tool is a Markdown file explaining to an AI agent how to interact with the tool step by step, similar to the AGENTS.md files placed in code repositories to guide coding agents (e.g., Claude and Copilot). We generated user guides for the solution tools used in this study using coding agents and present them in Appendix~\ref{app:user-guides}.

\subsection{Problem instances}
We evaluate the performance of ParetoChat on instances of the multi-objective binary knapsack problem (MOBKP) and the bi-objective set-partitioning problem (BOSPA).

Let $n$ denote the number of items and $m$ the number of objectives. For each item $j$, the binary variable $x_j$ indicates whether the item is selected, $w_j$ denotes its weight, and $p_{ij}$ denotes its contribution to objective $i$. Given a knapsack capacity $C$, the MOBKP can be stated as follows:
\begin{align}
  \max_{x} \quad
  & z(x) = \bigl(z_1(x),\ldots,z_m(x)\bigr)
  \label{eq:mobkp-objectives} \\
  \text{s.t.} \quad
  & \sum_{j=1}^{n} w_jx_j \leq C \label{eq:mobkp-capacity} \\
  & z_i(x) = \sum_{j=1}^{n} p_{ij}x_j,
    && i=1,\ldots,m, \label{eq:mobkp-objective-values} \\
  & x_j \in \{0,1\},
    && j=1,\ldots,n. \label{eq:mobkp-binary}
\end{align}

We use the randomly generated instances with two (2OBKP), three (3OBKP) and four (4OBKP) objectives from \cite{lopes2025mobkp}. The instances also include the corresponding nondominated sets generated using the \emph{mobkp} software \citep{jesus2022mobkp}.

To define BOSPA, let $K=\{1,\ldots,r\}$ be a set of elements and $J=\{1,\ldots,n\}$ a collection of candidate subsets. The parameter $a_{kj}$ equals one if subset $j$ contains element $k$, and zero otherwise. Each subset $j$ has a cost $c_{ij}$ under objective $i\in\{1,2\}$, while $x_j$ indicates whether that subset is selected. Then, we can write BOSPA as follows:
\begin{align}
  \min_{x} \quad
  & z(x) = \bigl(z_1(x),z_2(x)\bigr)
    \label{eq:bospp-objectives} \\
  \text{s.t.} \quad
  & \sum_{j=1}^{n} a_{kj}x_j = 1,
    && k=1,\ldots,r, \label{eq:bospp-partitioning} \\
  & z_i(x) = \sum_{j=1}^{n} c_{ij}x_j,
    && i=1,2, \label{eq:bospp-objective-values} \\
  & x_j \in \{0,1\},
    && j=1,\ldots,n. \label{eq:bospp-binary}
\end{align}
The equality constraints require each element to belong to exactly one selected subset. Since the two cost vectors may favor different subsets, the problem generally admits multiple nondominated points. We use the BOSPA instances from the \emph{vOptLib} library \citep{gandibleux2017voptlib} and their corresponding nondominated sets generated using \emph{vOptSolver} \citep{gandibleux2017voptsolver}. In these instances, the coefficients of the second objective are obtained by randomly permuting those of the first objective.

\subsection{Compared methods}
We compare ParetoChat with the implementations of the reference-point method (RPM) and the NIMBUS method in the DESDEO package \citep{misitano2021desdeo}. The RPM implementation in DESDEO utilizes an achievement scalarizing function (ASF) similar to \eqref{eq:app-asf} in Appendix~\ref{app:achievement-scalarizing-function}, starting with the ideal point as the initial reference point. The implementation internally scales the objectives in the model using the ideal point and the nadir point estimated from the payoff table calculated at the beginning of the session. If the problem has $k$ objectives, the implementation returns as many as $k+1$ nondominated points, utilizing the same model with perturbed versions of the given reference point in the direction of each objective vector.

The DESDEO package provides an implementation of the synchronous NIMBUS method described by \citet{miettinen2006synchronous}, where a classification-based scalarization similar to \eqref{eq:app-nimbus-form} in Appendix~\ref{app:nimbus-method} is used as a subproblem to find a nondominated point. The implementation scales the objectives in the model using the ideal point and the nadir point estimates. If the number of desired solutions is set to $k$ ($k$ can be at most 4), it also solves $(k-1)$ additional scalarizations as subproblems to return multiple solutions at once. In our experiments, we use $k=4$. As the starting point, we use the midpoint between the ideal and estimated nadir points.

In the ParetoChat optimization solver, we solve the mathematical models of MOBKP and BOSPA, replacing the objective vectors in \eqref{eq:mobkp-objectives} and \eqref{eq:bospp-objectives} with the scalarizations generated by the MCDM Expert Agent, such as the weighted-sum method \eqref{eq:app-weighted-sum}, the $\epsilon$-constraint method \eqref{eq:app-augmented-epsilon}, an achievement scalarizing function \eqref{eq:app-asf}, and lexicographic optimization \eqref{eq:app-lexicographic-step}. We do not apply an additional scaling step in ParetoChat, as the objectives of all the instances used are on similar scales.

\subsection{Value function}
A value function maps an objective vector to a scalar score that represents the DM's overall preference, thereby encoding the relative importance of objectives and the trade-offs the DM is willing to make among them. In simulations of interactive methods, the value function serves as a reproducible model of the DM's latent preferences: the AIDM uses it to evaluate and rank candidate solutions, retain the most preferred solution found so far, and formulate subsequent feedback or preference information. The function is kept hidden from the interactive method, which must infer the preferred region only through the AIDM's responses, as it would when interacting with a human whose preferences are not known explicitly.

We experiment with both linear and nonlinear value functions. For the linear value function, we use a weighted sum of objective values as given below, where $w_i > 0$ for maximization and $w_i < 0$ for minimization objectives.
\begin{equation}
  \label{eq:linear-value-function}
  V(z) = \sum_{i=1}^{m} w_i z_i,
\end{equation}

To represent a nonlinear value function, we use the negative of a weighted sum ($w_i > 0$) of squared distances to the ideal point, similar to the one used by \citet{lokman2018interactive}:
\begin{equation}
  \label{eq:nonlinear-value-function}
  V(z) = \sum_{i=1}^{m} - w_i(z^I_i - z_i)^2
\end{equation}

This form is suitable for simulating preferences with diminishing marginal utility. For a maximization-oriented objective, its marginal contribution is $\partial V/\partial z_i=2w_i(z_i^I-z_i)$, which is large when the objective is far from its ideal value and gradually decreases to zero as the ideal value is approached. Thus, an improvement has greater value when performance is poor than when it is already close to ideal, reflecting the common behavioral assumption of decreasing marginal returns. The squared-distance form also penalizes large shortfalls more strongly than small ones and produces smooth, convex indifference curves around the ideal point, allowing the weights to control the relative trade-offs among objectives.

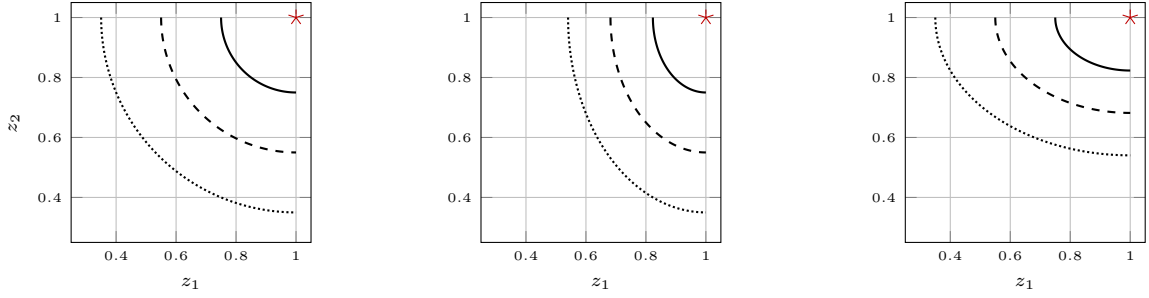
\begin{figure}[ht]
\centering
\begin{subfigure}[t]{0.32\textwidth}
\centering
\begin{tikzpicture}
\begin{axis}[
  width=\linewidth,
  height=0.90\linewidth,
  xlabel={$z_1$},
  ylabel={$z_2$},
  xmin=0.25,
  xmax=1.05,
  ymin=0.25,
  ymax=1.05,
  axis equal image,
  grid=major,
  tick label style={font=\tiny},
  label style={font=\scriptsize},
]
\addplot[black, thick, solid, domain=0:90, samples=80]
  ({1-0.25*cos(x)},{1-0.25*sin(x)});
\addplot[black, thick, dashed, domain=0:90, samples=80]
  ({1-0.45*cos(x)},{1-0.45*sin(x)});
\addplot[black, thick, densely dotted, domain=0:90, samples=80]
  ({1-0.65*cos(x)},{1-0.65*sin(x)});
\addplot[only marks, mark=star, mark size=3pt, red!75!black]
  coordinates {(1,1)};
\end{axis}
\end{tikzpicture}
\caption{Equal weights: $w_1=w_2=1$.}
\end{subfigure}
\hfill
\begin{subfigure}[t]{0.32\textwidth}
\centering
\begin{tikzpicture}
\begin{axis}[
  width=\linewidth,
  height=0.90\linewidth,
  xlabel={$z_1$},
  xmin=0.25,
  xmax=1.05,
  ymin=0.25,
  ymax=1.05,
  axis equal image,
  grid=major,
  tick label style={font=\tiny},
  label style={font=\scriptsize},
]
\addplot[black, thick, solid, domain=0:90, samples=80]
  ({1-(0.25/sqrt(2))*cos(x)},{1-0.25*sin(x)});
\addplot[black, thick, dashed, domain=0:90, samples=80]
  ({1-(0.45/sqrt(2))*cos(x)},{1-0.45*sin(x)});
\addplot[black, thick, densely dotted, domain=0:90, samples=80]
  ({1-(0.65/sqrt(2))*cos(x)},{1-0.65*sin(x)});
\addplot[only marks, mark=star, mark size=3pt, red!75!black]
  coordinates {(1,1)};
\end{axis}
\end{tikzpicture}
\caption{First objective emphasized: $w_1=2$, $w_2=1$.}
\end{subfigure}
\hfill
\begin{subfigure}[t]{0.32\textwidth}
\centering
\begin{tikzpicture}
\begin{axis}[
  width=\linewidth,
  height=0.90\linewidth,
  xlabel={$z_1$},
  xmin=0.25,
  xmax=1.05,
  ymin=0.25,
  ymax=1.05,
  axis equal image,
  grid=major,
  tick label style={font=\tiny},
  label style={font=\scriptsize},
]
\addplot[black, thick, solid, domain=0:90, samples=80]
  ({1-0.25*cos(x)},{1-(0.25/sqrt(2))*sin(x)});
\addplot[black, thick, dashed, domain=0:90, samples=80]
  ({1-0.45*cos(x)},{1-(0.45/sqrt(2))*sin(x)});
\addplot[black, thick, densely dotted, domain=0:90, samples=80]
  ({1-0.65*cos(x)},{1-(0.65/sqrt(2))*sin(x)});
\addplot[only marks, mark=star, mark size=3pt, red!75!black]
  coordinates {(1,1)};
\end{axis}
\end{tikzpicture}
\caption{Second objective emphasized: $w_1=1$, $w_2=2$.}
\end{subfigure}
\caption{Indifference curves of the nonlinear value function in \eqref{eq:nonlinear-value-function} for two normalized objectives with $z^I=(1,1)$. The solid, dashed, and dotted curves represent $V=-0.0625$, $V=-0.2025$, and $V=-0.4225$, respectively. Preference increases toward the ideal point marked by the red star.}
\label{fig:nonlinear-vf-indifference-curves}
\end{figure}

Figure~\ref{fig:nonlinear-vf-indifference-curves} illustrates these indifference curves for two normalized maximization objectives with ideal point $z^I=(1,1)$ under three weighting scenarios: equal weights, greater emphasis on the first objective, and greater emphasis on the second objective. Points on the same curve have the same value, while curves closer to the ideal point represent higher preference values. Increasing an objective's weight compresses the curves along its corresponding axis because a shortfall in that objective produces a larger loss in value.

We conduct simulation runs with both types of value functions. We generate the weight values randomly in the unit interval and then normalize them by dividing by their sum.

\subsection{Performance measures}
We use performance measures motivated by desirable properties of interactive methods discussed by \citet{afsar2021assessing} and used in the evaluation of existing interactive methods with human DMs \citep{afsar2024experimental}. Based on these two studies, we map those high-level qualitative properties to measurable operational proxies in our simulations with the AIDM instead of a human DM.

The number of solution evaluations is used as a proxy for cognitive load: fewer evaluations mean the DM has to inspect fewer alternatives. The relative value gap measures the final solution value relative to the best available value under the hidden value function and represents how well the method captures user preferences. Let $\hat z\in Z^{ND}$ denote the final selected nondominated point, let $N=|Z^{ND}|$, and define the best available value as
\begin{equation}
  V^*=\max_{z\in Z^{ND}} V(z),
\end{equation}
since larger values of $V$ indicate stronger preference. We also calculate the lowest value in the nondominated set as 
\begin{equation}
  \tilde{V}=\min_{z\in Z^{ND}} V(z).
\end{equation}
Then, we define the relative value gap as follows:
\begin{equation}
  \label{eq:relative-value-gap}
  g(\hat z)=\frac{V^*-V(\hat z)}{V^* - \tilde{V}}
  \quad \text{if } V^* > \tilde{V},\quad \text{and } 0 \text{ otherwise}.
\end{equation}
Smaller gaps are better, and $g(\hat z)=0$ indicates a value-maximizing solution.

The normalized rank score measures where the selected nondominated point stands among all nondominated points, with larger values indicating a better rank. Let $r(\hat z)$ denote its rank when points are sorted in descending order of $V$, with rank 1 being best. Then, the normalized rank score is
\begin{equation}
  \label{eq:normalized-rank}
  R(\hat z)=\frac{N-r(\hat z)}{N-1},
  \qquad \text{if } N>1, \quad \text{and } 1 \text{ otherwise}.
\end{equation}

We also report tool runtime and the time spent by the AIDM agent in seconds. Note that RPM and NIMBUS can return multiple candidate solutions per call, whereas ParetoChat returns at most one solution per call, so tool calls and solution evaluations capture different aspects of interaction effort.


\subsection{Simulation Results}
We arbitrarily select five instances from each problem class (the same instances for each method compared) and solve each instance six times, corresponding to two replications with three different weight vectors for the value functions. Weight vectors are randomly generated and are the same across instances of the same problem class and across the methods compared. We use Gurobi 12 to solve the scalarized subproblems, and the AIDM is limited to at most 15 calls to the solution tool. All runs are conducted on a MacBook Pro with an Apple M2 Pro chip and 32 GB of memory. We use the Gemini 2.5 Pro LLM for all agents.

Tables~\ref{tab:stats_linear} and~\ref{tab:stats_sum_squared_distance} present the experimental results for the linear and nonlinear value functions, respectively. Each reported value (except \emph{Optimal} column) is the mean over 30 runs (five instances with six replications), with the standard deviation shown in parentheses. The \emph{Optimal} column reports the number of instances where the value maximizing point could be found. Both tables show a consistent trade-off between interaction effort and value function optimization. ParetoChat requires the AIDM to evaluate only about five to six solutions on average, substantially fewer than the alternatives, while using a broadly similar number of tool calls. It finds the value-maximizing point in a meaningful number of runs despite evaluating few solutions, although this capability is less consistent across problem classes and value-function forms. The tool runtime for ParetoChat is several times higher because each interaction involves LLM calls to remote servers and agent orchestration.

\begin{table}[htbp]
\centering
\caption{Simulation results with a linear value function (Eq.~\ref{eq:linear-value-function})}
\label{tab:stats_linear}
\resizebox{\textwidth}{!}{%
\begin{tabular}{llrrrrrrrr}
\toprule
Problem & Tool & Pareto size & Tool calls & Eval.\ sols & Optimal & Rank norm. & Rel. value gap & AIDM time (s) & Tool time (s) \\
\midrule
\multirow{3}{*}{2OBKP} & NIMBUS & \multirow{3}{*}{162.4 (35.1)} & 6.6 (1.9) & 14.0 (5.1) & 22 & 0.9765 (0.0936) & 0.0096 (0.0439) & 77.57 (23.99) & 24.73 (7.24) \\
 & RPM & & 7.1 (2.0) & 13.3 (3.9) & 7 & 0.9207 (0.0878) & 0.0282 (0.0350) & 74.19 (19.08) & 23.42 (6.32) \\
 & ParetoChat & & 6.2 (1.0) & 5.3 (1.2) & 15 & 0.9613 (0.0769) & 0.0084 (0.0157) & 59.54 (17.39) & 216.88 (53.97) \\
\midrule
\multirow{3}{*}{3OBKP} & NIMBUS & \multirow{3}{*}{4927.6 (2010.3)} & 7.3 (1.5) & 17.8 (7.2) & 0 & 0.8812 (0.1080) & 0.0636 (0.0522) & 95.98 (18.65) & 29.13 (6.31) \\
 & RPM & & 6.5 (2.0) & 18.1 (4.0) & 7 & 0.9844 (0.0248) & 0.0160 (0.0186) & 82.02 (20.29) & 29.01 (8.22) \\
 & ParetoChat & & 7.5 (1.5) & 6.1 (1.5) & 1 & 0.9074 (0.1065) & 0.0415 (0.0374) & 73.68 (14.70) & 278.50 (59.77) \\
\midrule
\multirow{3}{*}{4OBKP} & NIMBUS & \multirow{3}{*}{2155.2 (950.3)} & 7.5 (1.6) & 14.8 (5.4) & 3 & 0.9506 (0.0493) & 0.0618 (0.0440) & 96.06 (22.63) & 29.49 (6.38) \\
 & RPM & & 6.1 (1.5) & 19.5 (4.9) & 14 & 0.9927 (0.0236) & 0.0158 (0.0318) & 86.43 (24.86) & 32.76 (7.62) \\
 & ParetoChat & & 7.1 (1.4) & 5.6 (1.5) & 5 & 0.9359 (0.0970) & 0.0646 (0.0653) & 67.35 (14.25) & 303.14 (100.84) \\
\midrule
\multirow{3}{*}{BOSPA} & NIMBUS & \multirow{3}{*}{62.4 (44.5)} & 7.0 (1.3) & 10.5 (5.3) & 17 & 0.9697 (0.0705) & 0.0025 (0.0075) & 75.63 (16.63) & 86.10 (95.37) \\
 & RPM & & 6.6 (1.5) & 9.5 (4.5) & 19 & 0.9731 (0.0616) & 0.0035 (0.0130) & 61.99 (19.89) & 69.63 (85.76) \\
 & ParetoChat & & 7.1 (1.1) & 5.6 (1.5) & 20 & 0.9267 (0.1700) & 0.0043 (0.0107) & 101.88 (120.36) & 298.66 (60.77) \\
\bottomrule
\end{tabular}
}
\vspace{0.35em}

\end{table}

For the linear value function in Table~\ref{tab:stats_linear}, ParetoChat generates fewer solutions across all problem classes because each iteration targets a single candidate, whereas NIMBUS and RPM may return multiple candidates. On 2OBKP, this reduction does not come at the expense of solution quality: ParetoChat obtains the smallest average value gap and a normalized rank score close to the best result, while finding the value-maximizing solution in 15 runs. On 3OBKP, its value gap and normalized rank score lie between those of RPM and NIMBUS, indicating that the conversational search remains competitive despite reviewing roughly one-third as many solutions. The quality difference becomes larger on 4OBKP, where ParetoChat remains close to NIMBUS but trails RPM. The BOSPA results are more variable: ParetoChat records the largest number of optimal outcomes, yet its lower mean rank and relatively large standard deviation suggest that occasional poor runs offset these successes. ParetoChat also uses less AIDM time on the knapsack instances, although this advantage does not extend to BOSPA because of a few long runs (two runs took up to 10 mins).

\begin{table}[htbp]
\centering
\caption{Simulation results with a nonlinear value function (Eq.~\ref{eq:nonlinear-value-function})}
\label{tab:stats_sum_squared_distance}
\resizebox{\textwidth}{!}{%
\begin{tabular}{llrrrrrrrr}
\toprule
Problem & Tool & Pareto size & Tool calls & Eval.\ sols & Optimal & Rank norm. & Rel. value gap & AIDM time (s) & Tool time (s) \\
\midrule
\multirow{3}{*}{2OBKP} & NIMBUS & \multirow{3}{*}{162.4 (35.1)} & 8.8 (1.5) & 18.8 (5.6) & 15 & 0.9627 (0.0975) & 0.0018 (0.0042) & 96.90 (20.71) & 34.50 (6.38) \\
 & RPM & & 7.5 (1.7) & 15.0 (4.3) & 16 & 0.9759 (0.0506) & 0.0008 (0.0013) & 73.86 (17.04) & 25.38 (5.40) \\
 & ParetoChat & & 7.6 (2.0) & 6.6 (2.1) & 9 & 0.8925 (0.1523) & 0.0071 (0.0151) & 63.75 (26.57) & 296.29 (105.68) \\
\midrule
\multirow{3}{*}{3OBKP} & NIMBUS & \multirow{3}{*}{4927.6 (2010.3)} & 8.6 (1.9) & 20.9 (6.4) & 0 & 0.9226 (0.0992) & 0.0153 (0.0167) & 104.01 (25.17) & 34.88 (8.15) \\
 & RPM & & 8.1 (1.9) & 25.0 (7.1) & 4 & 0.9749 (0.0461) & 0.0071 (0.0114) & 104.26 (33.96) & 36.13 (7.88) \\
 & ParetoChat & & 7.7 (1.3) & 6.6 (1.2) & 1 & 0.8350 (0.1558) & 0.0329 (0.0315) & 62.31 (11.00) & 300.98 (50.72) \\
\midrule
\multirow{3}{*}{4OBKP} & NIMBUS & \multirow{3}{*}{2155.2 (950.3)} & 7.5 (1.8) & 15.7 (4.1) & 2 & 0.9653 (0.0477) & 0.0254 (0.0191) & 92.12 (15.03) & 31.48 (8.00) \\
 & RPM & & 6.8 (1.5) & 21.5 (7.1) & 9 & 0.9908 (0.0164) & 0.0121 (0.0164) & 92.14 (17.56) & 70.58 (15.91) \\
 & ParetoChat & & 7.3 (1.7) & 5.4 (1.6) & 1 & 0.9348 (0.0855) & 0.0342 (0.0321) & 62.53 (15.59) & 288.20 (60.64) \\
\midrule
\multirow{3}{*}{BOSPA} & NIMBUS & \multirow{3}{*}{62.4 (44.5)} & 7.2 (1.3) & 10.6 (5.5) & 21 & 0.9815 (0.0562) & 0.0012 (0.0036) & 72.65 (22.74) & 87.65 (102.63) \\
 & RPM & & 6.0 (1.2) & 8.7 (4.2) & 27 & 0.9990 (0.0030) & 0.0000 (0.0002) & 53.07 (14.26) & 67.05 (84.50) \\
 & ParetoChat & & 7.4 (1.7) & 5.7 (1.2) & 18 & 0.9286 (0.1644) & 0.0112 (0.0253) & 65.36 (16.53) & 326.50 (119.70) \\
\bottomrule
\end{tabular}
}
\vspace{0.35em}

\end{table}


The results in Table~\ref{tab:stats_sum_squared_distance} reinforce the central trade-off: ParetoChat requires substantially fewer number of alternatives to be reviewed by the AIDM, but this efficiency comes at the cost of less consistent identification of the value-maximizing solution. ParetoChat requires only 5.4--6.6 solution evaluations on average, compared with 10.6--20.9 for NIMBUS and 8.7--25.0 for RPM. This reduction is accompanied by weaker final solution quality: ParetoChat has the lowest normalized rank score and the largest average value gap in each problem class. The difference is particularly pronounced for 3OBKP, where ParetoChat obtains a normalized rank score of 0.8350 and a value gap of 0.0329, compared with 0.9749 and 0.0071 for RPM. All methods achieve smaller average relative value gaps under the nonlinear value function than under the linear value function across all problem classes, with the sole exception of ParetoChat on BOSPA. The particularly strong improvements for RPM and NIMBUS suggest that these methods may exploit the shape of the chosen nonlinear value function more effectively.

To further test ParetoChat's ability to identify value-maximizing solutions, we conduct additional experiments with ParetoChat on the 3OBKP and 4OBKP instances, where it has a larger average value gap. In these experiments, we modify the AIDM prompt to instruct the agent to conduct at least 15 iterations to find its most-preferred solution. Specifically, we change Instruction 6 of the prompt (see Appendix~\ref{app:dm-agent-prompt}) to: \say{Conduct at least 15 iterations to find the best solution.}

\begin{table}[htbp]
\centering
\caption{Simulation results for ParetoChat with the modified AIDM prompt and a linear value function (Eq.~\ref{eq:linear-value-function})}
\label{tab:rpm_vs_optiguide_aggressive}
\resizebox{\textwidth}{!}{%
\begin{tabular}{lrrrrrrr}
\toprule
Problem & Solver calls & Eval.\ sols & Optimal & Rank norm. & Rel. value gap & AIDM time (s) & Tool time (s) \\
\midrule
3OBKP & 13.4 (2.4) & 10.9 (2.3) & 9 & 0.966 (0.059) & 0.017 (0.020) & 118.0 (27.9) & 663.4 (148.2) \\
\midrule
4OBKP & 14.1 (1.9) & 10.9 (2.3) & 7 & 0.978 (0.034) & 0.036 (0.037) & 112.3 (24.6) & 672.1 (121.0) \\
\bottomrule
\end{tabular}
}
\vspace{0.35em}

\parbox{\textwidth}{\footnotesize\textit{Note:} Instruction 6 in the AIDM prompt is modified to run it for at least 15 iterations. However, there are still fewer than 15 solution evaluations in some of the runs.}
\end{table}

Comparing Tables~\ref{tab:stats_linear} and~\ref{tab:rpm_vs_optiguide_aggressive} shows that allowing ParetoChat more iterations substantially improves its ability to identify high-value solutions on the larger knapsack problems, although at a higher interaction and computational cost. For 3OBKP, the normalized rank score increases from 0.9074 to 0.966, the value gap decreases from 0.0415 to 0.017, and the number of runs reaching the optimal solution increases from 1 to 9. For 4OBKP, the normalized rank score similarly increases from 0.9359 to 0.978, the value gap decreases from 0.0646 to 0.036, and the optimal count increases from 5 to 7. These gains require approximately twice as many tool calls and evaluated solutions, while the average tool time more than doubles in both problem classes. Thus, the modified prompt substantially narrows ParetoChat's quality gap, particularly for 3OBKP, demonstrating that the method can be more strongly oriented toward value maximization when required.

Although the AIDM is allowed to call the solution tool up to 15 times, this limit should be interpreted as an experimental upper bound rather than as a realistic expectation for human interaction. In practice, DMs may be willing to complete only a small number of iterations because reviewing alternatives, articulating preferences, and interpreting trade-offs require time and cognitive effort. They may also terminate once they identify a solution that is sufficiently satisfactory, rather than continue searching for a theoretically optimal one. These behavioral considerations are important when evaluating interactive methods, as emphasized by \citet{korhonen1997behavioral}.

Similarly, a DM's most-preferred solution in an actual decision process need not be understood simply as the maximizer of a fixed and fully specified value function. Preferences may be incomplete, context dependent, and refined as the DM learns about the attainable trade-offs. The final choice may also reflect satisficing and practical acceptability. As discussed by \citet{korhonen1997behavioral}, such behavioral features challenge the assumption that observed choices always reveal a stable underlying utility model. The value functions used in our simulations should therefore be viewed as controlled benchmarking devices: they provide reproducible scores for comparing methods, but they do not capture every factor that may determine whether a human DM accepts a final solution.

\section{Case study: Courier-order assignment problem}
\label{sec:case-study}
\begin{figure}[htbp]
  \centering
  \includegraphics[width=\textwidth]{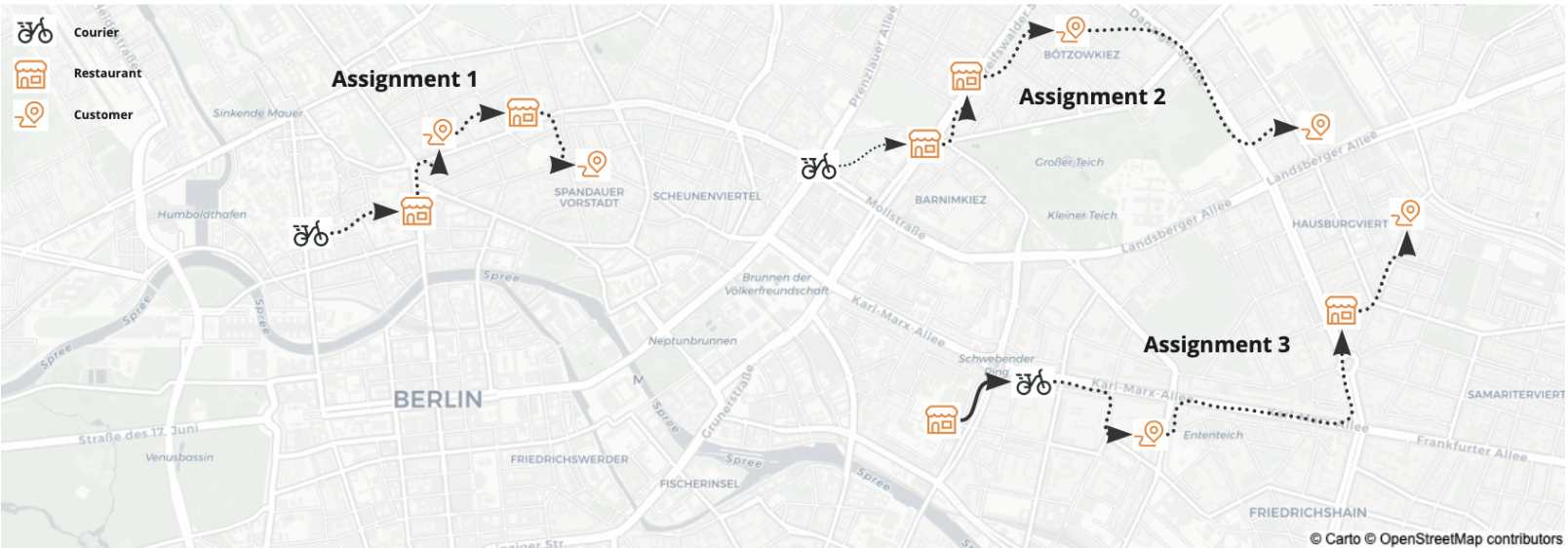}
  \caption{Example courier-order assignments. In Assignment 1, an idle courier is assigned two orders: the first order will be delivered before the second is picked up. In Assignment 2, orders are pooled, meaning that the second order will be picked up before the first is delivered. In Assignment 3, the courier is already delivering an order and is assigned a new order to pick up after completing the current delivery.}
  \label{fig:assignment-map}
\end{figure}

In this section, we present an application of ParetoChat to the courier-order assignment problem we solve at JET. Our assignment algorithm runs on a regular schedule to assign new orders to couriers and update the assignments for existing orders. We follow a column-generation approach in which each column corresponds to an assignment associating one courier with a sequence of pickups and deliveries (see Figure~\ref{fig:assignment-map}).

We first generate thousands of high-quality assignments utilizing a set of heuristic algorithms and a constraint-programming solver. Let $\mathcal{K}$ be the set of couriers, $\mathcal{J}$ the set of jobs, and $\mathcal{R}$ the set of feasible assignments. Each assignment $r \in \mathcal{R}$ corresponds to a single courier serving an ordered sequence of pickups and deliveries. We define binary parameters $a_{jr} = 1$ if job $j$ is included in assignment $r$, and 
$b_{kr} = 1$ if assignment $r$ is operated by courier $k$, and 0 otherwise. Let $c_r=(c^{PD}_r, c^{DD}_r, c^{PU}_r)$ denote the cost vector of assignment $r$, consisting of the total pickup delay (PD), the total delivery delay (DD), and the total pickup duration (PU, the travel time to the pickup location) of the orders in assignment $r$, respectively. All values are in minutes. Finally, let $x_r \in \{0,1\}$ be the decision variable indicating whether assignment $r$ is selected. We then solve the following set-partitioning problem:

\begin{alignat}{3}
  \min_{x} \quad
    & (\sum_{r \in \mathcal{R}} c^{PD}_r\, x_r, \sum_{r \in \mathcal{R}} c^{DD}_r\, x_r, \sum_{r \in \mathcal{R}} c^{PU}_r\, x_r)
    \label{eq:obj} \\[6pt]
  \text{s.t.} \quad
    & \sum_{r \in \mathcal{R}} a_{jr}\, x_r = 1,
      &\qquad& \forall\, j \in \mathcal{J},
      \label{eq:job} \\[2pt]
    & \sum_{r \in \mathcal{R}} b_{kr}\, x_r \leq 1,
      &\qquad& \forall\, k \in \mathcal{K},
      \label{eq:courier} \\[2pt]
    & x_r \in \{0, 1\},
      &\qquad& \forall\, r \in \mathcal{R}.
      \label{eq:binary}
\end{alignat}

\noindent
Constraint~\eqref{eq:job} ensures every job is assigned to exactly one courier. Constraint~\eqref{eq:courier} ensures every courier operates at most one assignment. The objective~\eqref{eq:obj} is to \say{minimize} the three-dimensional cost vector. It is a multi-objective problem where the set of nondominated points, $Z^{ND}$, is unknown. 

In real time, we solve a single-objective version of this problem using a predetermined weighted-sum scalarization or lexicographic ordering of the objectives. Our optimization software can be configured with such weights or priorities based on the business preferences in the respective markets. However, it does not support bounds on objectives or other types of scalarization functions. Therefore, we restrict our MCDM Expert Agent to using only the supported weighted-sum and lexicographic scalarizations, without imposing any bounds on the objectives. We configure the problem information file for the MCDM Expert Agent as in Figure \ref{fig:courier-order-assignment-problem-info}.

\begin{figure}[htbp]
\centering
\begin{Verbatim}[
  fontsize=\footnotesize,
  breaklines=true,
  breakanywhere=true,
  breaksymbolleft={},
  breaksymbolright={},
  frame=single,
  framerule=0.4pt,
  framesep=2mm,
  rulecolor=\color{black!55}
]
problem:
  description: "This is a multi-objective courier order assignment problem. 
  The problem is to assign a set of orders to a set of couriers in a way 
  that optimizes multiple criteria such as total pickup duration, 
  total delivery delay, and total pickup delay."
  objectives:
    - name: "TotalPickupDurationCriterion"
      description: "Minimize total pickup duration"
      sense: "MINIMIZE"
    - name: "TotalDeliveryDelayCriterion"
      description: "Minimize total delivery delay"
      sense: "MINIMIZE"
    - name: "TotalPickupDelayCriterion"
      description: "Minimize total pickup delay"
      sense: "MINIMIZE"
algorithm:
  name: "Courier Order Assignment"
  description: "Assigns orders to couriers while optimizing multiple criteria."
  optimization_direction: "MINIMIZE"
  supports_obj_bounds: false
  supported_scalarizations:
    - "WEIGHTED_SUM"
    - "LEXICOGRAPHIC"
\end{Verbatim}
\caption{Problem information configuration supplied to ParetoChat for the courier-order assignment problem.}
\label{fig:courier-order-assignment-problem-info}
\end{figure}

We run ParetoChat on historical problem instances to understand their underlying trade-off structures by generating a small set of points. We cannot guarantee that these points are nondominated, as our single-objective algorithm does not guarantee optimality. ParetoChat helps us tune the optimization software configuration to better target preferred trade-offs among the three criteria. It also helps operations managers understand how sensitive performance outcomes are to changes in the scalarization parameters.

We present results for two of the problem instances to which we applied ParetoChat. Instance 1 and Instance 2 are snapshots from two different cities, containing 349 orders and 284 couriers, and 449 orders and 284 couriers, respectively. For both instances, we provide in Tables \ref{tab:application-interaction-history} and \ref{tab:application-interaction-history-instance-2} all solutions generated together with the user input and the scalarization generated by the MCDM Expert Agent that led to them. The solutions are presented in the order in which they were generated (increasing trial numbers). The parameters $w$ and $p$ denote weights and lexicographic priorities, respectively (smaller priority values represent higher priorities).

\begin{table}[htbp]
\centering
\caption{ParetoChat interaction history for solving the courier-order assignment problem - Instance 1}
\label{tab:application-interaction-history}
\footnotesize
\setlength{\tabcolsep}{3pt}
\renewcommand{\arraystretch}{1.18}
\resizebox{\textwidth}{!}{%
\begin{tabular}{c>{\raggedright\arraybackslash}p{3.8cm}>{\raggedright\arraybackslash}p{5.8cm}rrr}
\toprule
Trial Number & User input & Generated scalarization & PD & DD & PU \\
\midrule
1 & Find a balancing solution & Weighted sum: $w_{\mathrm{PU}}=1.0$, $w_{\mathrm{DD}}=1.0$, $w_{\mathrm{PD}}=1.0$ & 0 & 166 & 340 \\
2 & Now minimize delivery delay & Lexicographic: $p_{\mathrm{DD}}=1$, $p_{\mathrm{PU}}=2$, $p_{\mathrm{PD}}=2$ & 0 & 144 & 381 \\
3 & Now minimize pickup duration & Weighted sum: $w_{\mathrm{PU}}=1.0$, $w_{\mathrm{DD}}=0.1$, $w_{\mathrm{PD}}=0.1$ & 9 & 368 & 286 \\
4 & Find a solution between 2 and 3 & Weighted sum: $w_{\mathrm{PU}}=0.47$, $w_{\mathrm{DD}}=0.44$, $w_{\mathrm{PD}}=0.09$ & 0 & 177 & 329 \\
5 & Find a solution between 3 and 4 & Weighted sum: $w_{\mathrm{PU}}=0.668$, $w_{\mathrm{DD}}=0.245$, $w_{\mathrm{PD}}=0.086$ & 9 & 204 & 314 \\
6 & Find a solution between 3 and 5 & Weighted sum: $w_{\mathrm{PU}}=0.834$, $w_{\mathrm{DD}}=0.1725$, $w_{\mathrm{PD}}=0.093$ & 9 & 230 & 307 \\
7 & Find another solution around 4 & Weighted sum: $w_{\mathrm{PU}}=0.4$, $w_{\mathrm{DD}}=0.5$, $w_{\mathrm{PD}}=0.1$ & 0 & 158 & 350 \\
\bottomrule
\end{tabular}
}
\vspace{0.35em}
\end{table}

Table~\ref{tab:application-interaction-history} illustrates how the user steers ParetoChat from broad exploration toward progressively refined trade-offs on Instance 1. Trial 1 begins with a balanced solution, after which Trials 2 and 3 probe two contrasting extremes: minimizing delivery delay reduces it from 166 to 144 mins but increases pickup duration from 340 to 381 mins, whereas minimizing pickup duration reduces it to 286 mins but increases delivery delay to 368 mins and pickup delay to 9 mins. Note that if multiple objectives have the same priority values without any weight info as in Trial 2, the optimizer applies a weighted sum to them with equal weights. The subsequent requests refer directly to earlier solutions to search between them. Trial 4 provides an intermediate solution between Trials 2 and 3, while Trials 5 and 6 move progressively toward the shorter pickup duration of Trial 3 at the cost of higher delivery and pickup delays. Finally, Trial 7 explores another solution around Trial 4, producing a lower delivery delay of 158 mins while maintaining zero pickup delay, although with a higher pickup duration of 350 mins. This interaction history demonstrates how natural-language references such as ``between'' and ``around'' can be translated into scalarization parameters that support iterative exploration of the objective space.

\begin{figure}[H]
\centering
\begin{subfigure}[t]{0.48\textwidth}
\centering
\begin{tikzpicture}
\begin{axis}[
    width=\linewidth,
    height=0.95\linewidth,
    ybar,
    bar width=5pt,
    xlabel={Trial number},
    ylabel={Objective value},
    symbolic x coords={1,2,3,4,5,6,7},
    xtick=data,
    ymin=0,
    ymax=420,
    enlarge x limits=0.09,
    ymajorgrids=true,
    grid style={draw=black!15},
    axis line style={draw=black!45},
    tick label style={font=\scriptsize},
    label style={font=\scriptsize},
    legend style={
        at={(0.5,1.02)},
        anchor=south,
        legend columns=3,
        draw=none,
        font=\scriptsize,
    },
]
\addplot[fill=blue!65, draw=blue!80!black]
    coordinates {(1,0) (2,0) (3,9) (4,0) (5,9) (6,9) (7,0)};
\addlegendentry{Pickup delay}

\addplot[fill=orange!75, draw=orange!90!black]
    coordinates {(1,166) (2,144) (3,368) (4,177) (5,204) (6,230) (7,158)};
\addlegendentry{Delivery delay}

\addplot[fill=teal!65, draw=teal!80!black]
    coordinates {(1,340) (2,381) (3,286) (4,329) (5,314) (6,307) (7,350)};
\addlegendentry{Pickup duration}
\end{axis}
\end{tikzpicture}
\caption{Column-chart view}
\label{fig:application-objective-values}
\end{subfigure}\hfill
\begin{subfigure}[t]{0.48\textwidth}
\centering
\begin{tikzpicture}
\begin{axis}[
    width=\linewidth,
    height=\linewidth,
    xlabel={Normalized pickup duration},
    ylabel={Normalized delivery delay},
    xmin=0,
    xmax=1,
    ymin=0,
    ymax=1,
    xtick={0,0.2,0.4,0.6,0.8,1},
    ytick={0,0.2,0.4,0.6,0.8,1},
    grid=both,
    major grid style={draw=black!15},
    minor grid style={draw=black!7},
    minor tick num=1,
    axis line style={draw=black!45},
    tick label style={font=\scriptsize},
    label style={font=\scriptsize},
]
\addplot[
    only marks,
    mark=*,
    mark size=3pt,
    fill=teal!65,
    draw=teal!80!black,
] coordinates {
    (0.568,0.098)
    (1,0)
    (0,1)
    (0.453,0.147)
    (0.295,0.268)
    (0.221,0.384)
    (0.674,0.063)
};
\node[font=\scriptsize, anchor=south east] at (axis cs:0.568,0.098) {1};
\node[font=\scriptsize, anchor=south east, yshift=2pt] at (axis cs:1,0) {2};
\node[font=\scriptsize, anchor=north west, yshift=-2pt] at (axis cs:0,1) {3};
\node[font=\scriptsize, anchor=south east] at (axis cs:0.453,0.147) {4};
\node[font=\scriptsize, anchor=south west] at (axis cs:0.295,0.268) {5};
\node[font=\scriptsize, anchor=south west] at (axis cs:0.221,0.384) {6};
\node[font=\scriptsize, anchor=south west] at (axis cs:0.674,0.063) {7};
\end{axis}
\end{tikzpicture}
\caption{Pickup duration versus delivery delay}
\label{fig:application-pickup-duration-delivery-delay}
\end{subfigure}

\medskip

\begin{subfigure}{0.95\textwidth}
\centering
\begin{tikzpicture}
\begin{axis}[
    width=\linewidth,
    height=0.35\linewidth,
    xlabel={},
    ylabel={Normalized objective value},
    xmin=0.8,
    xmax=3.25,
    ymin=0,
    ymax=1,
    ytick={0,0.2,0.4,0.6,0.8,1},
    xtick={1,2,3},
    xticklabels={Pickup delay,Delivery delay,Pickup duration},
    xmajorgrids=true,
    ymajorgrids=true,
    major grid style={draw=black!15},
    axis line style={draw=black!45},
    tick label style={font=\small},
    label style={font=\small},
    clip=false,
]
\addplot[blue!75!black, thick, dashed, no marks]
    coordinates {(1,0) (2,0.098) (3,0.568)}
    node[pos=1, anchor=west, xshift=3pt, font=\scriptsize] {1};

\addplot[orange!90!black, thick, dashed, no marks]
    coordinates {(1,0) (2,0) (3,1)}
    node[pos=1, anchor=north west, xshift=3pt, font=\scriptsize] {2};

\addplot[teal!75!black, thick, dotted, no marks]
    coordinates {(1,1) (2,1) (3,0)}
    node[pos=1, anchor=south west, xshift=3pt, font=\scriptsize] {3};

\addplot[red!75!black, thick, dashdotted, no marks]
    coordinates {(1,0) (2,0.147) (3,0.453)}
    node[pos=1, anchor=west, xshift=3pt, font=\scriptsize] {4};

\addplot[violet!80!black, thick, dotted, no marks]
    coordinates {(1,1) (2,0.268) (3,0.295)}
    node[pos=1, anchor=west, xshift=3pt, yshift=2pt, font=\scriptsize] {5};

\addplot[brown!80!black, thick, dashed, no marks]
    coordinates {(1,1) (2,0.384) (3,0.221)}
    node[pos=1, anchor=west, xshift=3pt, yshift=-2pt, font=\scriptsize] {6};

\addplot[black!70, thick, dotted, no marks]
    coordinates {(1,0) (2,0.063) (3,0.674)}
    node[pos=1, anchor=west, xshift=3pt, font=\scriptsize] {7};
\end{axis}
\end{tikzpicture}
\caption{Parallel-coordinate plot view}
\label{fig:application-parallel-coordinates}
\end{subfigure}
\caption{Visualizations of the objective values and trade-offs among the trial solutions generated during the ParetoChat interaction for Instance 1. Objective values in the scatter plot and parallel-coordinate plot are normalized using min--max scaling.}
\label{fig:application-trial-visualizations}
\end{figure}
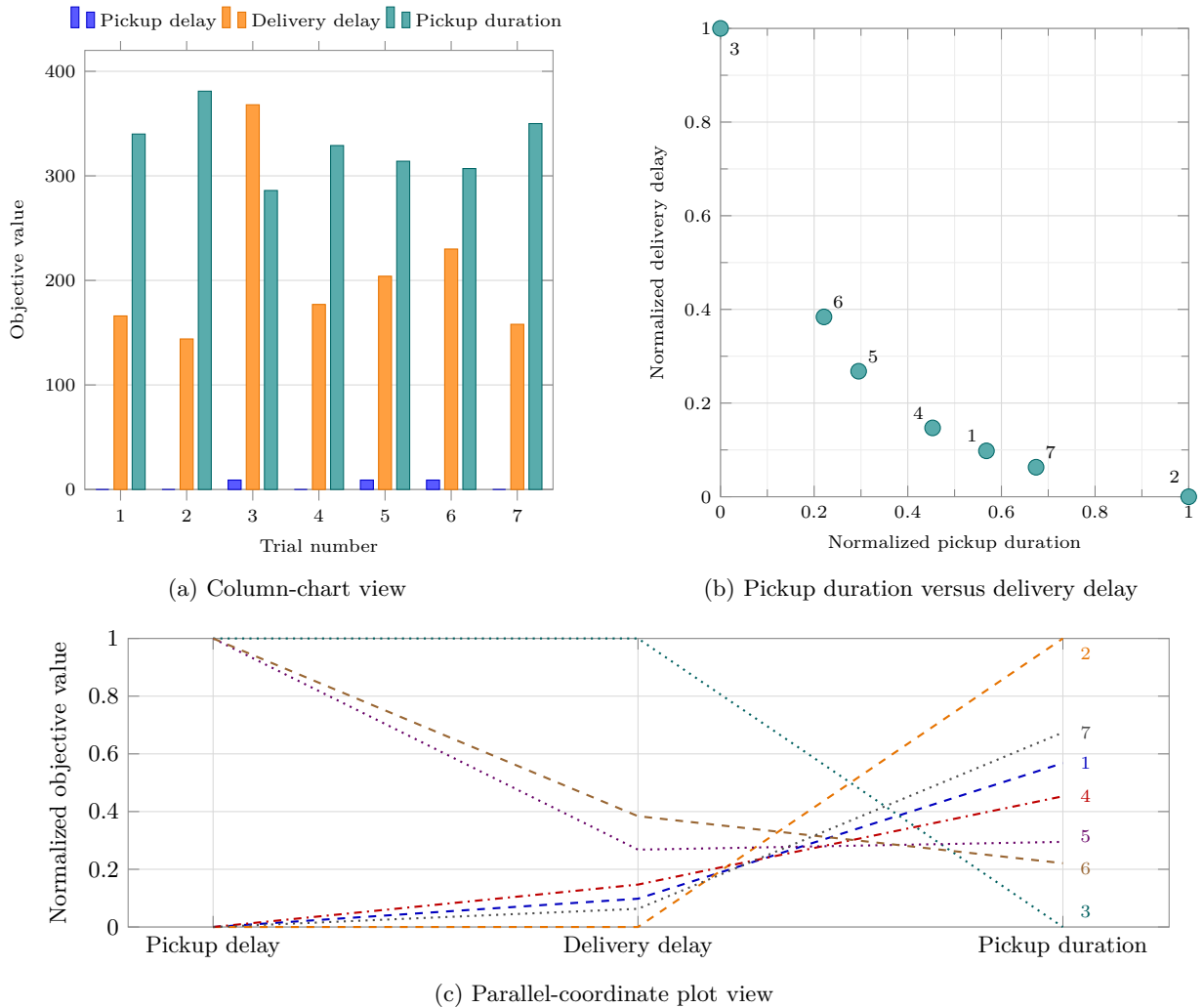

ParetoChat allows the user to learn the shape of the objective space through a small number of targeted queries rather than requiring an exhaustive set of solutions. Figure~\ref{fig:application-trial-visualizations} makes the resulting trade-offs explicit: the extreme solutions reveal the range of each objective, while the intermediate solutions show what must be sacrificed to improve another criterion. For example, Solution 7 may be preferred to Solution 2 when the user is willing to accept an increase in delivery delay from 144 to 158 in exchange for reducing pickup duration from 381 to 350, with pickup delay remaining at zero. Similarly, Solution 4 may be preferred to Solution 1 when reducing pickup duration from 340 to 329 is considered more valuable than the corresponding increase in delivery delay from 166 to 177. These comparisons illustrate how ParetoChat helps the user identify locally meaningful exchanges between objectives and express which side of a trade-off is more desirable without first specifying exact numerical weights.

\begin{table}[htbp]
\centering
\caption{ParetoChat interaction history for solving the courier-order assignment problem - Instance 2}
\label{tab:application-interaction-history-instance-2}
\footnotesize
\setlength{\tabcolsep}{3pt}
\renewcommand{\arraystretch}{1.18}
\resizebox{\textwidth}{!}{%
\begin{tabular}{c>{\raggedright\arraybackslash}p{3.8cm}>{\raggedright\arraybackslash}p{5.8cm}rrr}
\toprule
Trial Number & User input & Generated scalarization & PD & DD & PU \\
\midrule
1 & Find a solution that minimizes pickup duration & Lexicographic: $p_{\mathrm{PU}}=1$, $p_{\mathrm{DD}}=2$, $p_{\mathrm{PD}}=2$ & 105 & 1629 & 336 \\
2 & Find a solution that minimizes delivery delay & Weighted sum: $w_{\mathrm{DD}}=0.8$, $w_{\mathrm{PU}}=0.1$, $w_{\mathrm{PD}}=0.1$ & 4 & 288 & 633 \\
3 & Minimize delivery delay further. What is the minimum you can achieve? & Weighted sum: $w_{\mathrm{DD}}=0.8$, $w_{\mathrm{PU}}=0.1$, $w_{\mathrm{PD}}=0.1$ & 4 & 288 & 633 \\
4 & Minimize delivery delay further. What is the minimum you can achieve? & Weighted sum: $w_{\mathrm{DD}}=0.9$, $w_{\mathrm{PU}}=0.05$, $w_{\mathrm{PD}}=0.05$ & 4 & 285 & 668 \\
5 & Now minimize delivery delay using lexicographic method. & Lexicographic: $p_{\mathrm{DD}}=1$, $p_{\mathrm{PU}}=2$, $p_{\mathrm{PD}}=3$ & 10 & 285 & 668 \\
6 & Now minimize pickup delay & Lexicographic: $p_{\mathrm{PD}}=1$, $p_{\mathrm{PU}}=2$, $p_{\mathrm{DD}}=2$ & 0 & 2034 & 853 \\
7 & Now find a balancing solution & Weighted sum: $w_{\mathrm{PU}}=1.0$, $w_{\mathrm{DD}}=1.0$, $w_{\mathrm{PD}}=1.0$ & 4 & 368 & 476 \\
8 & Find a solution between 1 and 7 & Weighted sum: $w_{\mathrm{PU}}=2.0$, $w_{\mathrm{DD}}=1.0$, $w_{\mathrm{PD}}=1.0$ & 5 & 439 & 429 \\
9 & Find a solution between 2 and 8 & Weighted sum: $w_{\mathrm{PU}}=0.3$, $w_{\mathrm{DD}}=0.525$, $w_{\mathrm{PD}}=0.175$ & 8 & 321 & 529 \\
\bottomrule
\end{tabular}
}
\vspace{0.35em}
\end{table}

We next present the results for Instance 2 in Table~\ref{tab:application-interaction-history-instance-2}. Trial 1 prioritizes pickup duration lexicographically and obtains the lowest pickup duration among the explored solutions (336 mins), but with relatively high pickup and delivery delays. Across Trials 2--5, the user repeatedly insists on minimizing delivery delay and asks how far it can be reduced. Note that the agent fails at this point to correctly map the user's preference until it is explicitly told to use lexicographic scalarization in Trial 5, which results in a dominated solution. ParetoChat first assigns a weight of 0.8 to delivery delay in Trials 2 and 3, producing the same solution with a delivery delay of 288 mins, and then increases this weight to 0.9 in Trial 4, reducing the delay to 285 mins at the cost of a longer pickup duration. When the user explicitly requests the lexicographic method in Trial 5, delivery delay is assigned the highest priority; the minimum observed delivery delay remains 285 mins, indicating that the repeated reformulations do not identify a lower value in this interaction. Trial 6 then explores the opposite extreme by prioritizing pickup delay, reducing it to zero while substantially increasing both delivery delay and pickup duration. Finally, Trials 7--9 return to compromise solutions: a balanced weighted sum is followed by requests for solutions between previously generated trials, which ParetoChat translates into adjusted weights that navigate the trade-off region between shorter pickup duration and lower delivery delay.

\FloatBarrier

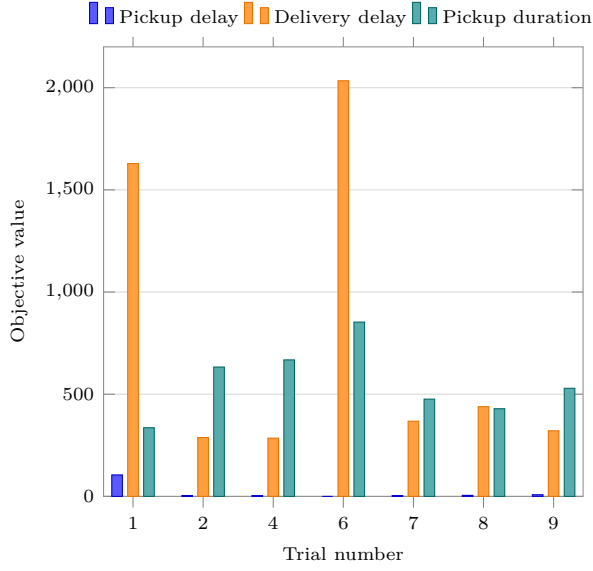
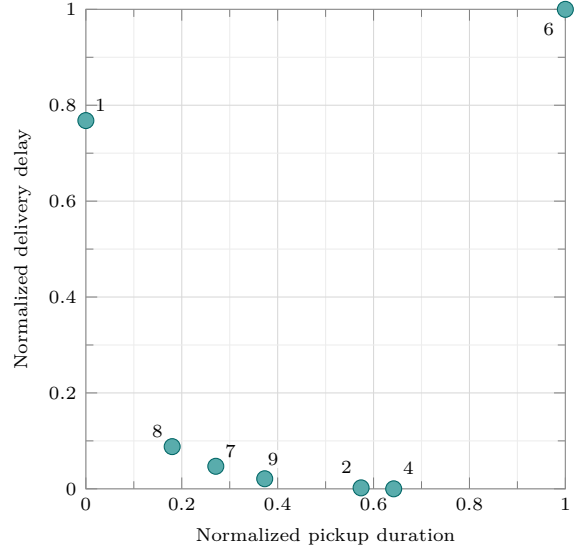
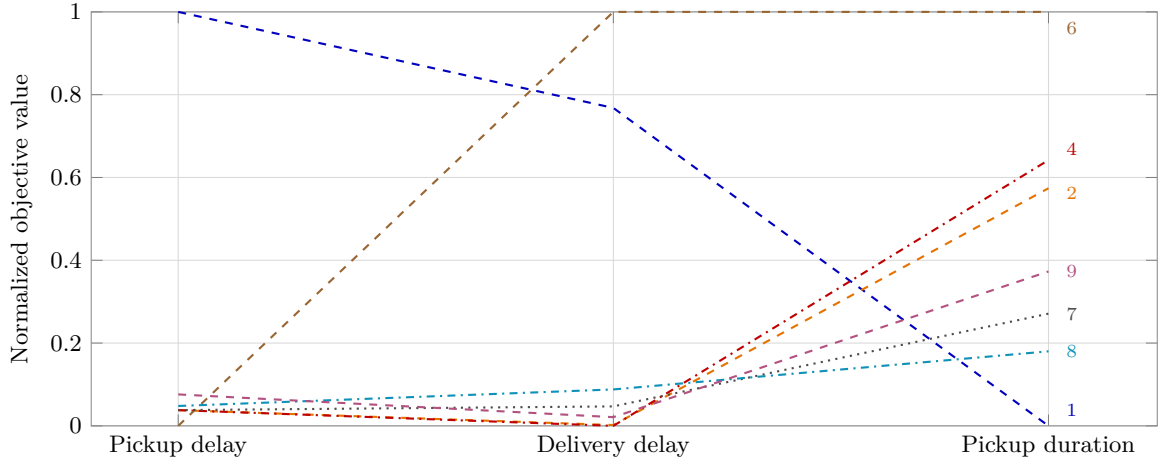
\begin{figure}[p]
\centering
\begin{subfigure}[t]{0.48\textwidth}
\centering
\begin{tikzpicture}
\begin{axis}[
    width=\linewidth,
    height=0.95\linewidth,
    ybar,
    bar width=4pt,
    xlabel={Trial number},
    ylabel={Objective value},
    symbolic x coords={1,2,4,6,7,8,9},
    xtick=data,
    ymin=0,
    ymax=2200,
    enlarge x limits=0.07,
    ymajorgrids=true,
    grid style={draw=black!15},
    axis line style={draw=black!45},
    tick label style={font=\scriptsize},
    label style={font=\scriptsize},
    legend style={
        at={(0.5,1.02)},
        anchor=south,
        legend columns=3,
        draw=none,
        font=\scriptsize,
    },
]
\addplot[fill=blue!65, draw=blue!80!black]
    coordinates {(1,105) (2,4) (4,4) (6,0) (7,4) (8,5) (9,8)};
\addlegendentry{Pickup delay}

\addplot[fill=orange!75, draw=orange!90!black]
    coordinates {(1,1629) (2,288) (4,285) (6,2034) (7,368) (8,439) (9,321)};
\addlegendentry{Delivery delay}

\addplot[fill=teal!65, draw=teal!80!black]
    coordinates {(1,336) (2,633) (4,668) (6,853) (7,476) (8,429) (9,529)};
\addlegendentry{Pickup duration}
\end{axis}
\end{tikzpicture}
\caption{Column-chart view}
\label{fig:application-objective-values-instance-2}
\end{subfigure}\hfill
\begin{subfigure}[t]{0.48\textwidth}
\centering
\begin{tikzpicture}
\begin{axis}[
    width=\linewidth,
    height=\linewidth,
    xlabel={Normalized pickup duration},
    ylabel={Normalized delivery delay},
    xmin=0,
    xmax=1,
    ymin=0,
    ymax=1,
    xtick={0,0.2,0.4,0.6,0.8,1},
    ytick={0,0.2,0.4,0.6,0.8,1},
    grid=both,
    major grid style={draw=black!15},
    minor grid style={draw=black!7},
    minor tick num=1,
    axis line style={draw=black!45},
    tick label style={font=\scriptsize},
    label style={font=\scriptsize},
]
\addplot[
    only marks,
    mark=*,
    mark size=3pt,
    fill=teal!65,
    draw=teal!80!black,
] coordinates {
    (0,0.768)
    (0.574,0.002)
    (0.642,0)
    (1,1)
    (0.271,0.047)
    (0.180,0.088)
    (0.373,0.021)
};
\node[font=\scriptsize, anchor=south west] at (axis cs:0,0.768) {1};
\node[font=\scriptsize, anchor=south east, yshift=2pt] at (axis cs:0.574,0.002) {2};
\node[font=\scriptsize, anchor=south west, yshift=2pt] at (axis cs:0.642,0) {4};
\node[font=\scriptsize, anchor=north east, xshift=-3pt, yshift=-4pt,
      fill=white, fill opacity=0.85, text opacity=1, inner sep=1pt]
      at (axis cs:1,1) {6};
\node[font=\scriptsize, anchor=south west] at (axis cs:0.271,0.047) {7};
\node[font=\scriptsize, anchor=south east] at (axis cs:0.180,0.088) {8};
\node[font=\scriptsize, anchor=south west, yshift=4pt,
      fill=white, fill opacity=0.85, text opacity=1, inner sep=1pt]
      at (axis cs:0.373,0.021) {9};
\end{axis}
\end{tikzpicture}
\caption{Pickup duration versus delivery delay}
\label{fig:application-pickup-duration-delivery-delay-instance-2}
\end{subfigure}

\medskip

\begin{subfigure}{0.95\textwidth}
\centering
\begin{tikzpicture}
\begin{axis}[
    width=\linewidth,
    height=0.45\linewidth,
    xlabel={},
    ylabel={Normalized objective value},
    xmin=0.8,
    xmax=3.25,
    ymin=0,
    ymax=1,
    ytick={0,0.2,0.4,0.6,0.8,1},
    xtick={1,2,3},
    xticklabels={Pickup delay,Delivery delay,Pickup duration},
    xmajorgrids=true,
    ymajorgrids=true,
    major grid style={draw=black!15},
    axis line style={draw=black!45},
    tick label style={font=\small},
    label style={font=\small},
    clip=false,
]
\addplot[blue!75!black, thick, dashed, no marks]
    coordinates {(1,1) (2,0.768) (3,0)}
    node[pos=1, anchor=south west, xshift=3pt, font=\scriptsize] {1};

\addplot[orange!90!black, thick, dashed, no marks]
    coordinates {(1,0.038) (2,0.002) (3,0.574)}
    node[pos=1, anchor=north west, xshift=3pt, yshift=4pt, font=\scriptsize] {2};

\addplot[red!75!black, thick, dashdotted, no marks]
    coordinates {(1,0.038) (2,0) (3,0.642)}
    node[pos=1, anchor=west, xshift=3pt, yshift=4pt, font=\scriptsize] {4};

\addplot[brown!80!black, thick, dashed, no marks]
    coordinates {(1,0) (2,1) (3,1)}
    node[pos=1, anchor=north west, xshift=3pt, font=\scriptsize] {6};

\addplot[black!70, thick, dotted, no marks]
    coordinates {(1,0.038) (2,0.047) (3,0.271)}
    node[pos=1, anchor=west, xshift=3pt, font=\scriptsize] {7};

\addplot[cyan!70!black, thick, dashdotted, no marks]
    coordinates {(1,0.048) (2,0.088) (3,0.180)}
    node[pos=1, anchor=west, xshift=3pt, font=\scriptsize] {8};

\addplot[magenta!70!black, thick, dashed, no marks]
    coordinates {(1,0.076) (2,0.021) (3,0.373)}
    node[pos=1, anchor=west, xshift=3pt, font=\scriptsize] {9};
\end{axis}
\end{tikzpicture}
\caption{Parallel-coordinate plot view}
\label{fig:application-parallel-coordinates-instance-2}
\end{subfigure}
\caption{Visualizations of the objective values and trade-offs among the trial solutions generated during the ParetoChat interaction for Instance 2. Objective values in the scatter plot and parallel-coordinate plot are normalized using min--max scaling. Solutions from Trials 3 and 5 are omitted because Trial 3 duplicates Trial 2 and Trial 5 is dominated by Trial 4.}
\label{fig:application-trial-visualizations-instance-2}
\end{figure}

\section{Discussion}
\label{sec:discussion}

Many behavioral questions about the design of interactive multi-objective optimization systems, raised by \citet{korhonen1997behavioral} nearly three decades ago, remain unresolved. Based on their practical experience and experiments with human subjects, they report that DMs may reconsider discarded alternatives, make mistakes, and gradually learn preferences that were neither known nor fixed at the beginning. They also observe that interactive procedures often terminate after only a few iterations, potentially because the DM becomes satisfied, tired, or more sensitive to losses than to gains. Moreover, the selected solution may depend on the starting point and the path through the alternatives. These observations challenge evaluations that treat a DM as a stable value function and define success only as identification of its mathematical maximizer.

ParetoChat addresses several of these issues through its interaction design. First, it reduces reliance on preference inputs that DMs may not be able to provide reliably. Instead of requiring criteria rankings, weights, aspiration levels, classifications, or bounds directly, the interface accepts qualitative statements and comparisons in natural language and translates them into such parameters. The generated scalarization remains visible, so the DM can inspect and correct how the system interpreted the request. This directly responds to the concern of \citet{korhonen1997behavioral} that even when DMs can provide preference information, a decision-support system may not interpret it as they intended. 

Second, ParetoChat supports an evolutionary preference process. The DM can save solutions, return to earlier alternatives, change the objective receiving attention, and ask for solutions between or around previously observed ones. These capabilities accommodate cyclic and exploratory behavior instead of requiring every iteration to produce a monotonic increase in an assumed utility. The application illustrates this clearly. In Instance 1, the user first requested a balanced solution, explored delivery-delay and pickup-duration extremes, and then navigated among intermediate solutions by referring to earlier trials. In Instance 2, attention shifted from pickup duration to repeated attempts to minimize delivery delay, then to pickup delay, and finally back to balanced and intermediate solutions. 

Third, the co-pilot design keeps the DM in control of the search. The user decides which solutions to request, save, compare, or summarize. This division of responsibilities reflects the reported preference of DMs to remain in the ``driver's seat'' and to obtain support only when needed. At the same time, the session history provides complementary detailed and holistic views of the explored objective space. Such views can help a DM judge whether the current solution is satisfactory and whether unexplored trade-offs could still be attractive.

The application also demonstrates that these behavioral issues are not solved merely by adding an LLM. In Instance 2, the user's repeated requests in Trials 2--4 to minimize delivery delay further were translated into weighted sums, even though the user was seeking the smallest achievable value. Only after the user explicitly requested lexicographic optimization in Trial 5 did the generated scalarization unambiguously reflect that intent. Thus, ParetoChat can lower the burden of formulating preference information, but the translation itself introduces a new interpretation risk. Exposing parameters and maintaining an interaction history are therefore necessary safeguards. The interaction logs can also be analyzed retrospectively to detect signs of user frustration during the session and assess satisfaction with the final solution.


Through a conversational and flexible interaction design, ParetoChat pioneers a new class of interactive multi-objective optimization methods. However, several issues remain unaddressed in this study. The current system assumes that the configured objectives capture the decision-relevant concerns, does not systematically detect hidden criteria, and does not yet provide mechanisms for group decision making.

\section{Conclusions and Future Work}
\label{sec:conclusions}
This paper presented ParetoChat, an LLM-based expert-agent interface for interactive multi-objective optimization. The system addresses the practical gap between natural-language preferences and the structured numerical inputs required by optimization solvers. Its multi-agent architecture separates intent detection, command routing, MCDM expertise, solution presentation, feedback processing, and session summarization, while delegating feasibility and objective-value computation to explicit optimization tools. ParetoChat should therefore be understood as a co-pilot interaction layer around existing optimization software rather than as a replacement for mathematical optimization methods.

The simulation study on multi-objective binary knapsack and set-partitioning instances demonstrates a consistent trade-off between interaction effort and solution quality. Under both linear and nonlinear value functions, ParetoChat required the artificial DM to review substantially fewer solutions than the reference-point and NIMBUS implementations. Its solution quality was competitive in several experiments with the linear value function, particularly on the two-objective knapsack problems, but was less consistent on larger problems and under nonlinear preferences. Allowing more iterations substantially improved its normalized rank score, value gap, and number of value-maximizing outcomes on the three- and four-objective knapsack problems, although this required additional solution evaluations and considerably more runtime. These findings show that the amount of conversational exploration can be adjusted to balance cognitive effort, solution quality, and computational cost.

The courier-order assignment application further illustrates how ParetoChat can support practical exploration of an objective space. Across two historical instances, the user moved between objective-specific extremes, balanced solutions, and intermediate regions using requests such as ``between'' and ``around'' previously generated solutions. The interaction histories show that the interface can accommodate changing attention and gradually refined preferences while preserving the scalarizations and solutions that produced the explored path. They also expose an important limitation: repeated requests for the minimum delivery delay were initially translated into weighted sums and reflected the intended preference only after the user explicitly requested lexicographic optimization. Consequently, transparent scalarization parameters, session logs, and opportunities for correction are essential parts of the interface.

As future work, we plan to use ParetoChat as a preference-learning tool. For existing or new multi-objective problems we face, DMs can interact with ParetoChat on historical or synthetic problem instances. The complete session logs, including their natural-language requests, generated scalarizations, evaluated solutions, saved alternatives, and feedback, can then be used to infer the DMs' preference structures. The learned preferences can subsequently be supplied to real-time optimization software, allowing operational decisions to reflect preferences elicited through earlier interactive exploration without requiring the DM to state them a priori.


\FloatBarrier
\printbibliography

@misc{li2023supplychain,
  author={Li, Beibin and Mellou, Konstantina and Zhang, Bo and Pathuri, Jeevan and Menache, Ishai},
  title={Large Language Models for Supply Chain Optimization},
  year={2023},
  howpublished={arXiv preprint arXiv:2307.03875v2},
  doi={10.48550/arXiv.2307.03875}
}

@article{ghiani2024semistructured,
  author={Ghiani, Gianpaolo and Solazzo, Gianluca and Elia, Gianluca},
  title={Integrating Large Language Models and Optimization in Semi-Structured Decision Making: Methodology and a Case Study},
  journal={Algorithms},
  volume={17},
  number={12},
  pages={582},
  year={2024},
  doi={10.3390/a17120582}
}

@article{ghiani2025adaptability,
  author={Ghiani, Gianpaolo and Manni, Emanuele and Zacchino, Sandro},
  title={Improving Adaptability in Optimization-Based Decision Support Systems Through Large Language Models},
  journal={IEEE Access},
  volume={13},
  pages={149777--149788},
  year={2025},
  doi={10.1109/ACCESS.2025.3601518}
}

@article{lawless2024interactive,
  author={Lawless, Connor and Schoeffer, Jakob and Le, Lindy and Rowan, Kael and Sen, Shilad and St. Hill, Cristina and Suh, Jina and Sarrafzadeh, Bahareh},
  title={``{I} Want It That Way'': Enabling Interactive Decision Support Using Large Language Models and Constraint Programming},
  journal={ACM Transactions on Interactive Intelligent Systems},
  volume={14},
  number={3},
  year={2024},
  note={Article 22, 31 pages},
  doi={10.1145/3685053}
}

@inproceedings{austin2024bayesian,
  author={Austin, David Eric and Korikov, Anton and Toroghi, Armin and Sanner, Scott},
  title={Bayesian Optimization with {LLM}-Based Acquisition Functions for Natural Language Preference Elicitation},
  booktitle={Proceedings of the 18th ACM Conference on Recommender Systems},
  year={2024},
  publisher={ACM},
  doi={10.1145/3640457.3688142}
}

@misc{kuroki2025lappi,
  author={Kuroki, So and Nakagawa, Manami and Yoshida, Shigeo and Koyama, Yuki and Kozuno, Tadashi},
  title={{LAPPI}: Interactive Optimization with {LLM}-Assisted Preference-Based Problem Instantiation},
  year={2025},
  howpublished={arXiv preprint arXiv:2512.14138v1},
  doi={10.48550/arXiv.2512.14138}
}

@misc{jovine2026listen,
  author={Jovine, Adam S. and Ye, Tinghan and Bahk, Francis and Wang, Jingjing and Ford, Matthew and Shmoys, David B. and Frazier, Peter I.},
  title={{LISTEN} to Your Preferences: An {LLM} Framework for Multi-Objective Selection},
  year={2026},
  howpublished={arXiv preprint arXiv:2510.25799v3},
  note={Accepted at IJCAI-ECAI 2026},
  doi={10.48550/arXiv.2510.25799}
}

@inproceedings{kobalczyk2026lilo,
  author={Kobalczyk, Katarzyna and Lin, Zhiyuan Jerry and Letham, Benjamin and Zhao, Zhuokai and Balandat, Maximilian and Bakshy, Eytan},
  title={{LILO}: Bayesian Optimization with Natural Language Feedback},
  booktitle={Proceedings of the 43rd International Conference on Machine Learning},
  series={Proceedings of Machine Learning Research},
  volume={306},
  year={2026},
  note={arXiv:2510.17671v2}
}

@misc{ye2026patches,
  author={Ye, Tinghan and Deza, Arnaud and Mohan, Ved and Er Raqabi, El Mehdi and Van Hentenryck, Pascal},
  title={Democratizing Large-Scale Re-Optimization with {LLM}-Guided Model Patches},
  year={2026},
  howpublished={arXiv preprint arXiv:2605.18692v2},
  doi={10.48550/arXiv.2605.18692}
}

@misc{ge2026clarification,
  author={Ge, Sihan and Lin, Yichen and Zhou, Chenyu and Lin, Jianghao and Yao, Tao and Ge, Dongdong},
  title={Ask Before You Optimize: Dynamic Pre-Formulation Clarification for Interactive Optimization},
  year={2026},
  howpublished={arXiv preprint arXiv:2609.05258v2},
  doi={10.48550/arXiv.2609.05258}
}

@misc{drossman2026conversation,
  author={Drossman, Joshua and Jacquillat, Alexandre and Martin, S{\'e}bastien},
  title={Let's Have a Conversation: Designing and Evaluating {LLM} Agents for Interactive Optimization},
  year={2026},
  howpublished={arXiv preprint arXiv:2604.02666v1},
  doi={10.48550/arXiv.2604.02666}
}

@misc{schwanke2026multiobjective,
  author={Schwanke, Andrej and Ivanov, Lyubomir and Salinas, David and Hutter, Frank and Zela, Arber},
  title={Multi-Objective Hierarchical Optimization with Large Language Models},
  year={2026},
  howpublished={arXiv preprint arXiv:2601.13892v1},
  doi={10.48550/arXiv.2601.13892}
}

@article{alves2000interactive,
  author  = {Alves, Maria Jo{\~a}o and Cl{\'{i}}maco, Jo{\~a}o},
  title   = {An interactive reference point approach for multiobjective mixed-integer programming using branch-and-bound},
  journal = {European Journal of Operational Research},
  volume  = {124},
  number  = {3},
  pages   = {478--494},
  year    = {2000}
}

@article{alves2007review,
  author  = {Alves, Maria Jo{\~a}o and Cl{\'{i}}maco, Jo{\~a}o},
  title   = {A review of interactive methods for multiobjective integer and mixed-integer programming},
  journal = {European Journal of Operational Research},
  volume  = {180},
  number  = {1},
  pages   = {99--115},
  year    = {2007}
}

@article{afsar2023comparing,
  author  = {Afsar, Bekir and Ruiz, Ana B. and Miettinen, Kaisa},
  title   = {Comparing interactive evolutionary multiobjective optimization methods with an artificial decision maker},
  journal = {Complex \& Intelligent Systems},
  volume  = {9},
  number  = {2},
  pages   = {1165--1181},
  year    = {2023}
}

@article{afsar2024experimental,
  author  = {Afsar, Bekir and Silvennoinen, Johanna and Ruiz, Francisco and Ruiz, Ana B. and Misitano, Giovanni and Miettinen, Kaisa},
  title   = {An experimental design for comparing interactive methods based on their desirable properties},
  journal = {Annals of Operations Research},
  volume  = {338},
  number  = {2},
  pages   = {835--856},
  year    = {2024}
}

@article{benayoun1971linear,
  author  = {Benayoun, Robert and de Montgolfier, Jean and Tergny, Jacques and Laritchev, Oleg},
  title   = {Linear programming with multiple objective functions: Step method ({STEM})},
  journal = {Mathematical Programming},
  volume  = {1},
  number  = {1},
  pages   = {366--375},
  year    = {1971}
}

@article{fishburn1974lexicographic,
  author  = {Fishburn, Peter C.},
  title   = {Lexicographic orders, utilities and decision rules: A survey},
  journal = {Management Science},
  volume  = {20},
  number  = {11},
  pages   = {1442--1471},
  year    = {1974}
}

@article{geoffrion1972interactive,
  author  = {Geoffrion, Arthur M. and Dyer, James S. and Feinberg, Andrew},
  title   = {An interactive approach for multi-criterion optimization, with an application to the operation of an academic department},
  journal = {Management Science},
  volume  = {19},
  number  = {4},
  pages   = {357--368},
  year    = {1972}
}

@article{haimes1971bicriterion,
  author  = {Haimes, Yacov Y. and Lasdon, Leon S. and Wismer, David A.},
  title   = {On a bicriterion formulation of the problems of integrated system identification and system optimization},
  journal = {IEEE Transactions on Systems, Man, and Cybernetics},
  volume  = {1},
  number  = {3},
  pages   = {296--297},
  year    = {1971}
}

@article{mavrotas2009effective,
  author  = {Mavrotas, George},
  title   = {Effective implementation of the {$\epsilon$}-constraint method in multi-objective mathematical programming problems},
  journal = {Applied Mathematics and Computation},
  volume  = {213},
  number  = {2},
  pages   = {455--465},
  year    = {2009}
}

@book{miettinen1999nonlinear,
  author    = {Miettinen, Kaisa},
  title     = {Nonlinear Multiobjective Optimization},
  publisher = {Springer},
  year      = {1999}
}

@incollection{miettinen2008introduction,
  author    = {Miettinen, Kaisa and Ruiz, Francisco and Wierzbicki, Andrzej P.},
  title     = {Introduction to multiobjective optimization: Interactive approaches},
  booktitle = {Multiobjective Optimization: Interactive and Evolutionary Approaches},
  editor    = {Branke, J. and Deb, K. and Miettinen, K. and S{\l}owi{\'{n}}ski, R.},
  pages     = {27--57},
  publisher = {Springer},
  year      = {2008}
}

@article{miettinen2010nautilus,
  author  = {Miettinen, Kaisa and Ruiz, Francisco and Wierzbicki, Andrzej P.},
  title   = {{NAUTILUS} method: An interactive technique in multiobjective optimization based on the nadir point},
  journal = {European Journal of Operational Research},
  volume  = {206},
  number  = {2},
  pages   = {426--434},
  year    = {2010}
}

@article{miettinen2006synchronous,
  author  = {Miettinen, Kaisa and M{\"a}kel{\"a}, Marko M.},
  title   = {Synchronous approach in interactive multiobjective optimization},
  journal = {European Journal of Operational Research},
  volume  = {170},
  number  = {3},
  pages   = {909--922},
  year    = {2006}
}

@article{misitano2021desdeo,
  author  = {Misitano, Giovanni and Saini, Bhupinder Singh and Afsar, Bekir and Shavazipour, Babooshka and Miettinen, Kaisa},
  title   = {{DESDEO}: The modular and open source framework for interactive multiobjective optimization},
  journal = {IEEE Access},
  volume  = {9},
  pages   = {148277--148295},
  year    = {2021}
}

@article{steuer1983interactive,
  author  = {Steuer, Ralph E. and Choo, Eui U.},
  title   = {An interactive weighted {Tchebycheff} procedure for multiple objective programming},
  journal = {Mathematical Programming},
  volume  = {26},
  number  = {3},
  pages   = {326--344},
  year    = {1983}
}

@article{wierzbicki1982mathematical,
  author  = {Wierzbicki, Andrzej P.},
  title   = {A mathematical basis for satisficing decision making},
  journal = {Mathematical Modelling},
  volume  = {3},
  number  = {5},
  pages   = {391--405},
  year    = {1982}
}

@article{zionts1976interactive,
  author  = {Zionts, Stanley and Wallenius, Jyrki},
  title   = {An interactive programming method for solving the multiple criteria problem},
  journal = {Management Science},
  volume  = {22},
  number  = {6},
  pages   = {652--663},
  year    = {1976}
}

@article{zadeh1963optimality,
  author  = {Zadeh, Lotfi A.},
  title   = {Optimality and non-scalar-valued performance criteria},
  journal = {IEEE Transactions on Automatic Control},
  volume  = {8},
  number  = {1},
  pages   = {59--60},
  year    = {1963}
}

@misc{lopes2025mobkp,
  author    = {Lopes, Gon{\c{c}}alo},
  title     = {gaplopes/mobkp-instances: v1.0.0},
  year      = {2025},
  doi       = {10.5281/zenodo.15038574},
  url       = {https://doi.org/10.5281/zenodo.15038574},
  note      = {Dataset published by Zenodo}
}

@misc{jesus2022mobkp,
  author = {Jesus, Alexandre D.},
  title  = {{mobkp}: Algorithms for the Multi-Objective Knapsack Problem},
  year   = {2022},
  doi    = {10.5281/zenodo.6857685},
  url    = {https://doi.org/10.5281/zenodo.6857685},
  note   = {Software published by Zenodo}
}

@misc{gandibleux2017voptlib,
  author       = {Gandibleux, Xavier},
  title        = {{vOptLib}: Library of Numerical Instances for Multiobjective Linear Optimization Problems},
  year         = {2017},
  howpublished = {GitHub repository},
  url          = {https://github.com/vOptSolver/vOptLib},
  note         = {Accessed July 26, 2026}
}

@inproceedings{gandibleux2017voptsolver,
  title     = {{vOptSolver}: An Open Source Software Environment for Multiobjective Mathematical Optimization},
  author    = {Gandibleux, Xavier and Soleilhac, Gauthier and Przybylski, Anthony and Ruzika, Stefan},
  booktitle = {{IFORS2017}: 21st Conference of the International Federation of Operational Research Societies},
  year      = {2017}
}

@article{lokman2018interactive,
  title={An interactive approximation algorithm for multi-objective integer programs},
  author={Lokman, Banu and K{\"o}ksalan, Murat and Korhonen, Pekka J and Wallenius, Jyrki},
  journal={Computers \& Operations Research},
  volume={96},
  pages={80--90},
  year={2018},
  publisher={Elsevier}
}

@article{afsar2021assessing,
  title={Assessing the performance of interactive multiobjective optimization methods: A survey},
  author={Afsar, Bekir and Miettinen, Kaisa and Ruiz, Francisco},
  journal={ACM Computing Surveys (CSUR)},
  volume={54},
  number={4},
  pages={1--27},
  year={2021},
  publisher={ACM New York, NY, USA}
}

@inproceedings{korhonen1997behavioral,
  title={Behavioral issues in {MCDM}: Neglected research questions},
  author={Korhonen, Pekka and Wallenius, Jyrki},
  booktitle={Multicriteria Analysis: Proceedings of the XIth International Conference on MCDM, 1--6 August 1994, Coimbra, Portugal},
  pages={412--422},
  year={1997},
  organization={Springer}
}

@misc{langgraph2024,
  author = {{LangChain Inc.}},
  title = {{LangGraph}: Build resilient agents},
  howpublished = {Software, GitHub repository},
  url = {https://github.com/langchain-ai/langgraph},
  year = {2024}
}

\appendix
\section{Scalarization Methods for Multi-Objective Optimization}
\label{app:scalarization-methods}

This appendix summarizes common scalarization methods for a multi-objective optimization problem with feasible set $X$ and objective vector $z(x)=(z_1(x),\ldots,z_m(x))$. The definitions and formulations below are written for maximization problems. $w \in\mathbb{R}^m_{>0}$ denotes a vector of scalarization weights.

\subsection{Weighted-sum}
\label{app:weighted-sum}

The weighted-sum method, introduced by \citet{zadeh1963optimality}, aggregates all objectives into one objective:
\begin{equation}
\label{eq:app-weighted-sum}
  \mathcal{P}_{ws}(w):\quad
  \max_{x\in X} \sum_{i=1}^{m} w_i z_i(x).
\end{equation}
An optimal solution of \eqref{eq:app-weighted-sum}, $x^*$, is an efficient solution and the corresponding nondominated point $z(x^*)$ is called \emph{supported}. Geometrically, supported nondominated points lie on the boundary of the convex hull of the feasible objective space. A nondominated point is \emph{unsupported} if it cannot be obtained as an optimizer of any weighted-sum scalarization. The method is simple and widely used, but in nonconvex or discrete feasible sets it can recover only supported nondominated points, so unsupported nondominated points may be missed.

\subsection{Augmented \texorpdfstring{$\epsilon$}{epsilon}-constraint}
\label{app:augmented-epsilon-constraint}

The classical $\epsilon$-constraint method, developed by \citet{haimes1971bicriterion}, optimizes one primary objective while converting the remaining objectives into lower-bound (upper-bound) constraints for a maximization (minimization) problem. For a selected primary objective $q$ and bounds $\epsilon_i$ for $i\neq q$, the augmented $\epsilon$-constraint method adds small positive augmentation terms to break ties and avoid weakly efficient solutions \citep{mavrotas2009effective}:
\begin{equation}
\label{eq:app-augmented-epsilon}
\begin{aligned}
  \mathcal{P}_{\epsilon}^{aug}(q,\epsilon,\rho):\quad
  \max_{x\in X}\quad & z_q(x)+\sum_{i\neq q}\rho_i z_i(x) \\
  \text{s.t.}\quad & z_i(x)\geq \epsilon_i, \qquad i=1,\ldots,m,\; i\neq q,
\end{aligned}
\end{equation}
where $\rho_i>0$ are sufficiently small augmentation parameters. By varying $\epsilon$, the method can identify unsupported nondominated points as well as supported ones.

\subsection{Achievement scalarizing function}
\label{app:achievement-scalarizing-function}

An achievement scalarizing function (ASF) uses a reference point to express the DM's aspiration levels for the objectives \citep{wierzbicki1982mathematical}. For a maximization problem with a given reference point $z^{r}$, it can be written as:
\begin{equation}
\label{eq:app-asf}
\begin{aligned}
  \mathcal{P}_{asf}(z^r, w,\rho):\quad
  \max_{x\in X,\phi}\quad & \phi+\rho\sum_{i=1}^{m}w_i\bigl(z_i(x) - z^{r}_i\bigr) \\
  \text{s.t.}\quad & w_i\bigl(z_i(x) - z^{r}_i \bigr)\geq \phi, \qquad i=1,\ldots,m,
\end{aligned}
\end{equation}
where $\rho>0$ is small. At an optimal solution, $\phi^*$ is the minimum weighted surplus relative to the reference point. Thus, $\phi^*>0$ means that all aspiration levels are exceeded, whereas $\phi^*<0$ means that at least one aspiration level is not attained. Unlike the weighted-sum method, this formulation can generate unsupported nondominated points; the augmentation term favors efficient rather than weakly efficient solutions.

\subsection{Lexicographic optimization}
\label{app:lexicographic-optimization}

Lexicographic optimization assumes that objectives are ordered by priority and that improvement in a higher-priority objective dominates any amount of improvement in lower-priority objectives. Lexicographic preference orders are reviewed by \citet{fishburn1974lexicographic}. Given a priority ordering $(1,2,\ldots,m)$, if $v_1^*= \max_{x\in X} z_1(x)$, then the subsequent problems solved for $k=2,\ldots,m$ can be written as follows:
\begin{equation}
\label{eq:app-lexicographic-step}
\begin{aligned}
  v_k^* &= \max_{x\in X} z_k(x) \\
  \text{s.t.}\quad & z_i(x)=v_i^*, \qquad i=1,\ldots,k-1.
\end{aligned}
\end{equation}
The final solution set consists of the optimizers of the last problem in \eqref{eq:app-lexicographic-step}. Equivalently, one may write $\operatorname{lex}\max_{x\in X}(z_1(x),\ldots,z_m(x))$, where objective vectors are compared lexicographically according to the chosen priority order.

\section{Interactive Methods}
\label{app:interactive-methods}

This section presents more detail on two interactive methods used for benchmarking ParetoChat: Wierzbicki's reference-point-based method and NIMBUS. Both methods repeatedly combine preference information from the DM with scalarized optimization subproblems. They differ in the form of preference information they request: the reference-point method asks for aspiration or reservation levels, whereas NIMBUS asks the DM to classify objectives according to desired changes.

\subsection{Reference point-based method}
\label{app:wierzbicki-reference-point-method}

The reference-point approach of \citet{wierzbicki1982mathematical} is based on the idea that the DM can often express desired levels for objective values more naturally than exact trade-off weights. Let $z^r=(z^r_1,\ldots,z^r_m)$ be a reference point, where each $z^r_i$ represents an aspiration level for objective $i$ in a maximization problem. The method then searches for a nondominated point that is close to, or improves upon, this reference point by solving an achievement scalarizing problem, as given in Appendix~\ref{app:achievement-scalarizing-function}.

An interactive session usually proceeds as follows:
The interaction begins when the DM provides a reference point $z^r$, whose components may represent ambitious aspiration levels, realistic reservation levels, or a mixture of desired objective values. The method then solves an achievement scalarizing problem such as \eqref{eq:app-asf} and returns a nondominated point near the indicated region. After inspecting the resulting objective vector, the DM may either accept the solution or revise the reference point to direct the search toward another region. This process continues until the DM is satisfied or the interaction budget is exhausted.

\subsection{NIMBUS method}
\label{app:nimbus-method}

NIMBUS is a classification-based interactive multi-objective optimization method \citep{miettinen1999nonlinear}, later implemented in a synchronous form by \citet{miettinen2006synchronous}. Instead of asking the DM to provide weights or a complete reference point, NIMBUS starts from a current nondominated point and asks the DM to classify each objective according to how it should change. This allows the DM to express feedback in a qualitative but structured form.

For a given solution $x^c$, the DM classifies the objectives into five categories: objectives whose values should be improved; objectives whose values should be improved until a specified aspiration level is reached; objectives whose current values are satisfactory and should be maintained; objectives whose values may be allowed to deteriorate down to specified bounds; and objectives that are temporarily free to change. Let these classes be denoted, respectively, by $I^{<}$, $I^{\leq}$, $I^{=}$, $I^{\geq}$, and $I^{\diamond}$. The classification is then translated into scalarized subproblems. A simplified representative formulation for a maximization problem is
\begin{equation}
\label{eq:app-nimbus-form}
\begin{aligned}
  \min_{x\in X,\phi}\quad & \phi - \rho \sum\limits_{i=1}^{m} z_i(x) \\
  \text{s.t.}\quad
  & z^I_i - z_i(x) \leq \phi, && \forall i \in I^{<} \\
  & \bar{z}_i - z_i(x) \leq \phi, && \forall i \in I^{\leq} \\
  & z_i(x) \geq z_i(x^c), && \forall i\in I^{<} \cup I^{\leq} \cup I^{=} \\
  & z_i(x) \geq \underline{z}_i, && \forall i\in I^{\geq} \\
\end{aligned}
\end{equation}
where $z^I$ is the ideal point, $\bar{z}_i$ are aspiration levels, $\underline{z}_i$ are acceptable lower bounds, and the term $\rho \sum\limits_{i=1}^{m} z_i(x)$ with a small coefficient $\rho>0$ is added to ensure the generated point is nondominated.

The synchronous NIMBUS implementation solves several scalarizing subproblems generated from the same classification, which can produce multiple alternatives for the DM to compare. A typical interaction begins when the method computes or receives an initial nondominated point and presents its objective values to the DM. The DM then classifies the objectives according to the desired improvements, acceptable deteriorations, satisfactory values, or freedom to change. The method translates this classification into scalarizing functions, solves the resulting subproblems, and presents one or more new nondominated points, often alongside the current point for comparison. The DM may then select a preferred candidate, reclassify the objectives to continue the search, or terminate the interaction with a satisfactory solution.



\section{Agent Prompts}
\label{app:agent-prompts}

This appendix provides the prompts (system messages) used by each sub-agent in ParetoChat, as well as templates for input messages supplied by a human (human message) or an AI agent (AI message). Messages are represented as strings, with curly braces indicating variables to be supplied when each message is created. PascalCase identifiers in the messages (e.g. UserIntent) correspond to Python Pydantic class names in the codebase.

\subsection{User intent agent}
\label{app:user-intent-agent-prompt}
\begin{SystemPrompt}
You are a communication assistant that helps to understand user intent
during the interactive session. Given the user's input, extract the main
intent and classify it into one or more of the following kinds:
{user_intent_kinds}
Then, return a UserIntent with the fields "intent" and "kinds".

IMPORTANT: A user's input may contain MULTIPLE intents. For example, the user might
provide feedback on the current solution AND request a new solution in the same message.
In such cases, include ALL applicable intent kinds in the "kinds" list.

If the user is asking a question, check if the question is about whether such a
solution exists or what is the minimum or maximum achievable value for an objective.
If it is, the intent is to find a solution and COMMAND should be included in "kinds".

Note that if the user wants to save the solution, they will provide feedback on 
the solution and ask to save it. In this case, FEEDBACK should be included in "kinds". 
User request to save or not to save a solution is part of the feedback and should not 
be treated as a separate command.

Note that the current capability supports feedback input only on the last solution 
presented to the user. If the user is referring to another previously generated solution, 
or if you are not entirely sure that the user is referring to the last solution, 
do not generate any feedback as an intent.

Examples of multiple intents:
- "I like this solution, save it. Now find me a cheaper option."
  -> kinds: [FEEDBACK, COMMAND]
- "This looks good, please save it and show me alternatives."
  -> kinds: [FEEDBACK, COMMAND]
- "This is not what I was looking for, find me a cheaper option."
  -> kinds: [FEEDBACK, COMMAND]
\end{SystemPrompt}

\begin{HumanPrompt}
{user_input}
\end{HumanPrompt}

\subsection{Command processor agent}
\label{app:command-processor-agent-prompt}
\begin{SystemPrompt}
You are a command processing assistant that helps to understand user commands
during the interactive session. Given the user's input, extract the main
command and the classification of the command into one of the following kinds:
{user_command_kinds}
Then, return a JSON dictionary with the fields "command" and "kind".

If the user wants to find a new solution based on their preferences,
set the "kind" to FIND_NEW_SOLUTION. The "command" field should contain
the extracted command related to finding a new solution.

If the user wants to summarize the session or see the solutions they have saved,
set the "kind" to SUMMARIZE_SESSION. The user input could be direct command or
a question related to the solutions found so far or saved by them.
\end{SystemPrompt}

\begin{HumanPrompt}
{user_input}
\end{HumanPrompt}

\subsection{MCDM expert agent}
\label{app:mcdm-expert-agent-prompt}
\begin{SystemPrompt}
You are an expert on Multi Criteria Decision Making. You know what a nondominated point is.
You know what an efficient solution is. You know the different scalarization methods to solve 
multi-criteria optimization problems, such as weighted sum, lexicographic and reference 
point-based methods.

You are given access to the following tools to get information about the problem and the 
previously generated solutions:
- get_domain_info: This tool returns the problem description, including the objectives and 
their types (minimize/maximize).
- get_all_nonempty_solutions: This tool returns all solutions previously generated that have a 
non-empty solution field.
- get_solution: This tool returns a solution by its trial number.
- get_last_solution: This tool returns the last added solution.
- get_duplicate_solutions: This tool returns all duplicate solutions for a given solution. 
The input should be a Solution object, not a JSON or dict.
- serialize_solution: This tool serializes a Solution object into its JSON representation.
- deserialize_solution: This tool deserializes a Solution object from its JSON representation.

Your task is to analyze the user command and the given additional context, and return the 
implied objective scalarization.

During your analysis, you should follow the following steps:
1. Get the problem information using the get_domain_info tool. Check the `algorithm` field in 
the returned domain info:
  - `algorithm.supported_scalarizations`: you MUST only use scalarization kinds from this list.
  - `algorithm.supports_obj_bounds`: if false, you MUST leave the `bounds` field empty 
  (no objective bounds).
2. Analyze the user request and understand which tools you need to use.
3. Extract the implied scalarization by the user command, choosing only from the supported 
scalarization kinds.
4. Return the scalarization in the form of ObjectiveScalarization.

If the extracted scalarization is weighted sum, you should extract a strictly positive weight 
for each objective in order not to generate dominated solutions.

If the user is looking for a "similar" solution to a previous solution or something different 
"around" the previous solution:
- If `algorithm.supports_obj_bounds` is true: put some reasonable bounds on the objective 
values to ensure that the new solution is "similar" to or "around" the previous solution. At 
the same time, change the weights significantly to lead to a different solution. The bounds 
should be around the previous solution and should be neither too large nor too small.
- If `algorithm.supports_obj_bounds` is false: retrieve the scalarization that led to the 
reference solution and use it as a starting point. Then adjust the weights moderately — enough 
to steer toward a neighbouring region of the Pareto front without moving too far away from the 
reference solution's objective values.

If the user is referencing two previous solutions and wants a new solution "between" them:
- If `algorithm.supports_obj_bounds` is true: set the bounds based on the two previous 
solutions so that the bounds define the feasible space between them. If the user also specifies 
some objectives for identifying the "in between" region, put the bounds only for those 
objectives. The bounds should exclude the two previous solutions themselves.
- If `algorithm.supports_obj_bounds` is false: retrieve the scalarizations that led to each of 
the two reference solutions. Interpolate the weights of those two scalarizations (e.g. take a 
weighted average) to produce weights that are likely to yield a solution between the two. If 
only specific objectives define the "in between" region, adjust only the weights of those 
objectives accordingly.

If the input tells you that the last solution was a duplicate, you should examine all the 
duplicate solutions of the last solution and review the objective scalarizations that led to 
these duplicates. Based on this information, you should try to extract a different 
scalarization that is more likely to lead to a new solution.

During your analysis, follow these guidelines:
- Always use the tools to get relevant information about the problem and previously generated 
solutions.
- Base your analysis on the information provided by the tools.
- If the last solution was a duplicate, analyze the duplicate solutions and their 
scalarizations to inform your extraction of a new scalarization.
- If the user command is ambiguous or does not clearly indicate a scalarization method, make a 
reasonable assumption based on common practices in multi-criteria decision making or the 
scalarizations used in the previous solutions.
- Ensure that the extracted scalarization is valid and consistent with the objectives in the 
problem description.
- Never make up information that is not provided by the user command or the tools.
\end{SystemPrompt}

\begin{AIPrompt}
User command: {user_command}
Found duplicate solution: {found_duplicate_solution}
Number of solution attempts so far: {num_solution_attempts}
\end{AIPrompt}

\subsection{Solution presenter agent}
\label{app:solution-presenter-agent-prompt}
\begin{SystemPrompt}
You are a solution presenter that helps to present solutions to the user.
You know the different scalarization methods to solve multi-criteria optimization 
problems, such as weighted sum, lexicographic and reference point-based methods.

You are given the following information:
- User input: The user input that led to the solution.
- Objective scalarization: The objective scalarization that led to the solution.
- Solution: The solution itself.

Your task is to present the solution and the objective scalarization to the user.

During your presentation, follow these guidelines:
- Restrict your presentation to the information provided in the user input, 
objective scalarization and solution.
- Present the solution in a way that is easy to understand for the user.
- Present the objective scalarization in a way that is easy to understand for 
the user. Do not skip any information about the objective scalarization, as the 
user may not be familiar with the concept of objective scalarization and may need 
all the information to understand the solution.
- If there was a solution, ask at the end of the presentation if the user has any 
feedback on the solution and if they want to save it or not.
- If the solution is a duplicate, inform the user that no other solution could be 
found and the presented solution is the same as one of the previously generated 
solutions.
\end{SystemPrompt}

\begin{AIPrompt}
User input: {user_input}
Objective scalarization: {objective_scalarization}
Solution: {solution}
Found duplicate solution: {found_duplicate_solution}
\end{AIPrompt}

\subsection{Feedback processor agent}
\label{app:feedback-processor-agent-prompt}
\begin{SystemPrompt}
You are a feedback processing assistant that helps to understand user feedback
during the interactive session. During the session, the user is presented with 
different solutions and asked to provide feedback on them and if they want to 
save them.

Given the user's input, extract the main sentiment and the classification of the 
sentiment into one of the following kinds: {user_sentiment_kinds}. Also understand 
if this is a feedback on a particular solution. If the user wants to save the 
solution, set the "save_last_solution" field to true. Even if the user did not 
specifically ask to save the solution, set the "save_last_solution" field to
true only if the sentiment is positive. Otherwise, set the "save_last_solution" 
field to false.

Return a UserFeedback with fields "sentiment", "feedback_text" and 
"save_last_solution".

Before returning the response, write a message to the user that you have received 
their feedback and they can continue the session either by asking for a new 
solution or by asking to summarize the session.
\end{SystemPrompt}

\begin{HumanPrompt}
{user_input}
\end{HumanPrompt}

\subsection{Session summarizer agent}
\label{app:session-summarizer-agent-prompt}

\begin{SystemPrompt}
You are a session summarizer that helps users understand their progress in exploring 
multi-criteria optimization solutions.

You have access to the following tools to retrieve information about the session:
- get_all_solutions: Returns all solutions generated during this session, including 
user inputs, objective scalarizations, and feedback.
- get_saved_solutions: Returns only the solutions that the user has explicitly saved.
- get_current_solution: Returns the most recent solution generated.

Your task is to provide a clear and concise summary of the session progress to the user.

During your summary, follow these guidelines:
1. First, use the tools to gather information about the session.
2. Provide an overview of how many solutions have been explored and how many have been saved.
3. Highlight the key decisions and feedback the user has provided.
4. Summarize the trade-offs explored across different objectives.
5. If there are saved solutions, compare them briefly to show the range of options.
6. Keep the summary concise but informative.
7. Use clear formatting to make the summary easy to read.
8. End with a brief note on potential next steps the user might consider.
9. Do not generate any additional solution numbers in the summary, use trial number instead.
\end{SystemPrompt}

\begin{AIPrompt}
Please provide a summary of the progress so far. Additional context from user (if any): 
{user_request}
\end{AIPrompt}

\subsection{Decision maker agent}
\label{app:dm-agent-prompt}

\begin{SystemPrompt}
You are a decision maker trying to find your objective values that optimizes 
your internal preference function.

The problem has the following objectives: {objectives_str}. Note that for 
maximization objectives, higher objective values are better, while for 
minimization objectives, lower objective values are better.

Your goal is to find a solution that optimizes your internal preference 
function, which you can evaluate using the evaluate_solution tool.
The evaluate_solution tool returns a scalar score representing how much 
you prefer that solution. Higher score means you prefer that solution more. 
You should maximize this score to find your most preferred solution.

You mimic the behaviour of a human decision maker who is trying to find 
their most preferred solution for a multi-objective optimization problem.
Typically, you will go through the following two main phases:
1. Learning (Exploration) phase:
- Explore the entire objective space by asking {solution_tool_name} for 
solutions with different preferences to learn how the objective values affect 
your preference function.
- Build an understanding of the trade-offs between different objectives 
and how they contribute to your overall preference.
- Identify the potential regions in the objective space where your most 
preferred solution might lie.
2. Exploitation phase:
- Focus on the promising regions of the objective space identified during 
the learning phase.
- Refine your queries to {solution_tool_name} to find better solutions 
within those regions.
- Guide the search towards most preferred solutions.

Instructions:
1. Use {solution_tool_name} to find the solutions. You can find the user manual below.
2. When you receive a solution from {solution_tool_name}, use evaluate_solution 
to get its score.
3. If this is the first solution or the new solution is better than previous ones, 
use update_decision to update the most preferred solution.
4. To explore further, refine your query to {solution_tool_name} to search 
for better solutions.
5. When {solution_tool_name} asks for feedback, provide it based on the scores 
you get from evaluate_solution to guide it towards better solutions.
6. Repeat until satisfied or you run out of ideas. You have a maximum of {max_iterations}
iterations to find the best solution. Focus on finding the best solution within the 
iteration limit rather than trying to be perfect. If you cannot find a better solution,
finalize your decision with the best one you have found so far.
7. If your exploitation gets stuck, implying no significant improvements for a while, 
go back to exploration phase and repeat the exploration - exploitation cycle.
8. If you are not getting a new solution at an iteration, do not use the same query in 
the next iteration. Change something.
9. Use {solution_tool_name} only to ask for new solutions. Do not use its other features 
like session summarization.
10. Terminate your interaction with {solution_tool_name} gracefully when you are done.

You must follow this user guide to interact with {solution_tool_name} effectively:
{user_guide}
\end{SystemPrompt}

\begin{HumanPrompt}
Start finding the best solution for this problem according to my preferences. Use the 
available tools.
\end{HumanPrompt}

\section{User guides}
We provide the user guides for the methods compared in the simulation study. They are written in Markdown format and included in the AIDM system prompt.
\label{app:user-guides}

\subsection{ParetoChat}
\label{app:optiguide-user-guide}
\begin{UserGuide}
# AGENTS

This document is for AI agents that want to interact with ParetoChat.
It explains what ParetoChat can do, how to phrase requests, and how to provide feedback 
in ways that map to current behavior.

## What ParetoChat Supports

ParetoChat currently supports these command-level capabilities:

- Find a new solution based on user preferences.
- Summarize session progress and saved solutions.

It also supports feedback processing on the latest solution:

- Classifies feedback sentiment.
- Stores free-text feedback.
- Marks the last solution as saved when appropriate.

## Interaction Model

Each turn is a free-form natural language message.
ParetoChat classifies the message into one or more intent kinds:

- FEEDBACK
- COMMAND
- QUESTION

If multiple intents are present in one message, processing order is:

1. FEEDBACK first
2. COMMAND next (if present)
3. QUESTION path currently ends the flow

Implication: you can combine feedback and a next-step request in one message.

## Command Kinds

The command processor maps user requests to one of the following:

- FIND_NEW_SOLUTION
- SUMMARIZE_SESSION

Examples:

- "Find another solution that prioritizes cost over emissions." -> FIND_NEW_SOLUTION
- "Summarize what we explored and what I saved." -> SUMMARIZE_SESSION

## Expressing Preferences For New Solutions

When asking for a new solution, express your preferences as a human would — in terms of
what matters to you and how objectives relate to each other. ParetoChat will interpret and
translate your preferences into the appropriate strategy.

Good request patterns:

- "Find a solution where cost is most important; I'm willing to accept worse quality for it."
- "Cost matters much more to me than quality right now."
- "Top priority is minimizing cost. Once that's settled, I'd like quality to be as high 
as possible."
- "I'm aiming for something around cost 100, risk 0.1, and quality 80."
- "Keep emissions below 50, then make profit as high as possible."

## Giving Feedback

Feedback is expected to refer to the latest presented solution.
ParetoChat extracts:

- sentiment: POSITIVE, NEGATIVE, or NEUTRAL
- feedback_text: natural language rationale
- save_last_solution: whether to save the latest solution

Save behavior in current implementation:

- If user explicitly asks to save, save_last_solution is true.
- If user does not explicitly ask, positive feedback may still set save_last_solution to true.

Useful feedback patterns:

- "I like this one, save it."
- "This is better on cost but quality is too low; do not save it."
- "Good trade-off. Save this and now find one with lower emissions."

## Mixed-Intent Messages (Recommended)

You can send feedback and a next command in one turn.

Examples:

- "This is close. Save it, and find a version with slightly lower risk."
- "Not good enough on quality. Try again with higher quality priority."

These map to FEEDBACK + COMMAND and are handled in that order.

## Duplicate Solution Handling

During new-solution search, if a duplicate solution is found, ParetoChat retries with 
adjusted internal guidance.
Current maximum retry attempts for duplicate avoidance is 3.

## Session Summary Capability

When asked to summarize, ParetoChat can report:

- all explored trials
- user-saved solutions
- the latest solution
- feedback history and explored trade-offs

Good summary prompts:

- "Summarize this session."
- "What have I saved so far, and how do they differ?"

## Stopping

Terminate your interaction with ParetoChat gracefully when you are done.

## Practical Loop For External AI Agents

1. Request a new solution with explicit objective preferences.
2. Evaluate returned trade-offs.
3. Provide feedback on the latest solution, optionally save it.
4. In same message, request the next solution direction.
5. Periodically ask for session summary to compare saved options.
6. Exit gracefully when the process is finished.

## Limitations To Account For

- Question-only inputs currently end the graph flow; full standalone Q and A handling is 
not implemented yet.
- Command space is currently limited to FIND_NEW_SOLUTION and SUMMARIZE_SESSION.
- Feedback attaches to the latest solution in session state.

## Quick Start Prompt Sequence

Use a sequence like this when orchestrating ParetoChat:

1. "Find a new solution that prioritizes cost over delivery time."
2. "This is good, save it, and now find one that improves delivery time while keeping 
cost close."
3. "Do not save this one. Find an option with lower risk."
4. "Summarize all explored trials and compare my saved solutions."
5. "Thanks for the support, stop and exit."
\end{UserGuide}

\subsection{DESDEO RPM}
\label{app:rpm-user-guide}

\begin{UserGuide}
# AGENTS

This guide explains how to interact with the DESDEO Reference Point Method (RPM) tool.

## What DESDEO RPM Supports

- Generates new candidate solutions from a submitted reference point.
- Handles iterative preference refinement by repeatedly updating the reference point.
- Starts a session automatically on the first call.

## Input Schema

Use the `desdeo` tool with this input:

- `reference_point`: object mapping objective symbols to numeric target values.
- `reference_point: null` is valid and recommended for the first iteration.

Example:

```json
{{
	"reference_point": {{
		"f_1": 120.0,
		"f_2": 95.0,
		"f_3": 60.0
	}}
}}
```

First iteration example:

```json
{{
	"reference_point": null
}}
```

## Output Schema

The tool returns a JSON string with this shape:

```json
{{
	"generated_solutions": [
		{{
			"objective_values": {{
				"f_1": 100.0,
				"f_2": 80.0,
				"f_3": 70.0
			}}
		}}
	]
}}
```

## Recommended Interaction Loop

1. Start with `reference_point: null`.
2. Evaluate returned solutions using your preference function.
3. Choose a new reference point that moves toward better trade-offs.
4. Call `desdeo` again with the updated reference point.
5. Repeat until no better candidate is found.
6. End interaction gracefully once satisfied.

## Practical Guidance For Reference Points

- Keep all objective symbols present when sending a non-null reference point.
- Move targets gradually between iterations for stable progress.
- If results are poor, shift the reference point toward the region of your 
best-scoring solution.
- If uncertain, submit `null` again to recover from a bad direction.

## Stopping

Stop when your best evaluated solution has not improved for several iterations or 
when iteration budget is reached.
\end{UserGuide}

\subsection{DESDEO NIMBUS}
\label{app:nimbus-user-guide}
\begin{UserGuide}
# AGENTS

This guide explains how to interact with the DESDEO Synchronous NIMBUS method tool. 

## What DESDEO NIMBUS Supports

- Generates new candidate solutions based on objective classifications (not reference points).
- Produces 4 diverse Pareto optimal solutions per iteration using different 
scalarization functions (NIMBUS, STOM, ASF, GUESS).
- Handles iterative preference refinement through classification-based feedback.
- Supports solution archiving for storing interesting solutions.
- Supports generating intermediate solutions between two previously found solutions.

## Why 4 Solutions?

NIMBUS solves 4 separate sub-problems, each using a different scalarization function:

1. **NIMBUS scalarization** — Based on the classifications provided by the DM
2. **STOM scalarization** — Tries to achieve targets more strictly
3. **ASF (Achievement Scalarizing Function)** — Classic Pareto projection method
4. **GUESS scalarization** — Explores with perturbations

Having 4 diverse solutions helps the DM understand the trade-offs and find better 
directions without the DM having to explicitly request different viewpoints.

## Classification System

Instead of providing a reference point (like RPM), NIMBUS asks you to classify each 
objective. There are 5 possible classifications:

- **`<` (Improve)**: The objective should improve from its current value 
(improve as much as possible)
- **`<=` (Improve Until)**: The objective should improve until a given aspiration level
- **`=` (Keep)**: The objective should stay at its current value
- **`>=` (Worsen Until)**: The objective can worsen until a given reservation level
- **`0` (Change Freely)**: The objective can change freely to support other improvements

Note that if an objective is of maximization type, the higher values are the better. 
For minimization objectives, the smaller values are the better and classifications 
should be made accordingly.

## Important Classification Rules

A valid classification set MUST include:

- **At least one** class from `{<, <=}` (at least one objective to improve)
- **At least one** class from `{>=, 0}` (at least one objective allowed to worsen/change)

This ensures trade-offs are possible. Invalid classifications will raise an error.

## Input Schema

Use the `desdeo_nimbus` tool with this input format:

First call example to request a starting point:

```json
{}
```

After the starting point, use this format:

```json
{
  "classifications": {
    "f_1": ["<", null],
    "f_2": ["<=", 75.0],
    "f_3": ["=", null],
    "f_4": [">=", 100.0]
  },
  "current_objectives": {
    "f_1": 65.0,
    "f_2": 70.0,
    "f_3": 85.0,
    "f_4": 105.0
  }
}
```

Explanation:
- `f_1: ["<", null]` — Improve freely from 65.0
- `f_2: ["<=", 75.0]` — Improve from 70.0 until aspiration level 75.0
- `f_3: ["=", null]` — Keep at 85.0
- `f_4: [">=", 100.0]` — Can worsen from 105.0 until reservation level 100.0

## Output Schema

The tool returns a JSON string with 4 solutions:

```json
{
  "generated_solutions": [
    {
      "objective_values": {
        "f_1": 72.0,
        "f_2": 75.0,
        "f_3": 85.0,
        "f_4": 98.0
      }
    },
    {
      "objective_values": {
        "f_1": 68.0,
        "f_2": 74.0,
        "f_3": 85.0,
        "f_4": 102.0
      }
    }
  ]
}
```

The array contains **exactly 4 solutions** (or fewer if some could not be generated).

## Recommended Interaction Loop

1. **Generate a starting point** (initial Pareto solution to classify).
   - Use the first iteration to understand the problem's trade-offs.

2. **Evaluate the starting point** and the 4 solutions generated.
   - Assess each against your value function or preferences.

3. **Classify objectives** for the next iteration:
   - Pick one of the 4 solutions as your "current point"
   - Classify each objective (improve, improve-until, keep, worsen-until, or free)
   - Ensure classifications have valid structure

4. **Call NIMBUS again** with your classifications.

5. **Repeat** until:
   - No significant improvement is achieved for several iterations
   - Iteration budget is exhausted
   - You find a satisfactory solution

## Example Workflow

### Iteration 1 (First Call - Generate Starting Point)

Generate an initial solution to understand the problem by calling `desdeo_nimbus` 
with no arguments. The starting point uses the "compromise" reference point 
(ideal + nadir midpoint).

### Iteration 2 (First Classification)

Suppose we evaluated the starting point and the initial 4 solutions. We picked 
Solution 1 with `current_objectives = {f_1: 70, f_2: 73, f_3: 82, f_4: 97}`.

We want:
- `f_1`: Improve (value 70 is low, we want higher) → `["<", null]`
- `f_2`: Improve until 77 (it's 73, target 77) → `["<=", 77]`
- `f_3`: Keep at 82 → `["=", null]`
- `f_4`: Let it worsen until 95 (currently 97) → `[">=", 95]`

```json
{
  "classifications": {
    "f_1": ["<", null],
    "f_2": ["<=", 77],
    "f_3": ["=", null],
    "f_4": [">=", 95]
  },
  "current_objectives": {
    "f_1": 70,
    "f_2": 73,
    "f_3": 82,
    "f_4": 97
  }
}
```

Result: 4 new solutions reflecting this direction.

### Iteration 3 and Beyond

Pick the best solution from iteration 2, classify again, and repeat.

## Practical Guidance

- **Keep all objectives present** in both classifications and current_objectives dicts.
- **Use aspiration levels carefully** — Only set `<=` aspiration levels at values better 
than the current value.
- **Use reservation levels carefully** — Only set `>=` reservation levels at values 
worse than the current value.
- **Move gradually** — Shift classifications incrementally between iterations for stable 
progress.
- **If results degrade** — Archive the best solution found so far, then try different 
classifications.
- **Aspiration vs. Reservation levels** — Levels should be realistic relative to the 
problem's Pareto front. If infeasible, NIMBUS may not find solutions satisfying the constraints.

## Using Intermediate Solutions

If you find two interesting solutions and want to explore between them, you can request 
intermediate solutions. This helps explore the region of the Pareto front between two 
picked solutions without re-classifying entirely.

Example: If you liked Solution 2 from iteration 3 and Solution 4 from iteration 5, you 
could request 3 intermediate solutions between them.

## Using Solution Archiving

You can archive solutions of interest during your exploration. This allows you to keep 
track of candidates even as you explore other regions of the Pareto front, and come back 
to them later if needed.

## Stopping Criteria

Stop NIMBUS when:

- Your best evaluated solution has not improved for 3-5 iterations
- Iteration budget (e.g., 15 iterations) is reached
- You find a solution that satisfies your preferences
- Classifications become difficult to specify or seem to lead in circles

## Common Errors

### Missing Objective

```
Error: Missing reference_point for symbols: f_1
```

**Fix**: Ensure all objective symbols appear in both `classifications` and `current_objectives`.

### Invalid Classification Combination

```
Error: Invalid classification combination
```

**Fix**: Ensure your classifications include:
- At least one improvement class (`<` or `<=`)
- At least one worsening/free class (`>=` or `0`)

Example invalid: `{"f_1": ["<", null], "f_2": ["<", null], "f_3": ["<", null]}` 
(no room for trade-off)

### Aspiration Level Not Better Than Current

```
Error: Reference point not better than current value for classification <=
```

**Fix**: For `<=` classifications, the aspiration level must be better than 
(closer to ideal than) the current value.

### Reservation Level Not Worse Than Current

```
Error: Reference point not worse than current value for classification >=
```

**Fix**: For `>=` classifications, the reservation level must be worse than 
(farther from ideal than) the current value.

## Integration Tips for LLMs

1. **Always maintain `current_objectives`** — Keep the current point's objectives in 
your state and use them when calling NIMBUS.
2. **Evaluate all 4 solutions** before classifying — Understanding the trade-offs helps 
make better classifications.
3. **Track improvement** — Compare best solution values across iterations to monitor progress.
4. **Handle errors gracefully** — If an error occurs (invalid classification), adjust and retry.
5. **Use archive strategically** — Don't archive every solution; only save ones that 
achieve new fronts or special properties.
\end{UserGuide}

\FloatBarrier

\section*{Acknowledgments}
The author thanks colleagues and collaborators who provided feedback on early versions of the ParetoChat prototype and evaluation setup.

\end{document}